%% file: main.tex
\documentclass[a4paper,fleqn]{cas-dc}

\usepackage[authoryear]{natbib}
\usepackage{graphicx}       

\usepackage{epstopdf}
\usepackage{color}
\usepackage[noend]{algpseudocode}
\usepackage{algorithmicx,algorithm}

\usepackage[utf8]{inputenc} 
\usepackage[T1]{fontenc}    
\usepackage{hyperref}       
\usepackage{url}            
\usepackage{booktabs}       
\usepackage{amsmath,amssymb,amsfonts}
\usepackage{nicefrac}       
\usepackage{lipsum}
\usepackage{fancyhdr}       
\usepackage{setspace}
\usepackage{stfloats}
\usepackage{float}
\usepackage{multicol}
\usepackage{array}
\usepackage{multirow}
\usepackage{enumitem}
\usepackage{cuted}
\usepackage{caption}
\usepackage{tabularx}
\usepackage{hhline}
\usepackage{threeparttable}
\usepackage{subcaption}
\usepackage{soul}
\usepackage{cases}
\usepackage{pgfplots}
\pgfplotsset{compat=1.18}
\usepackage[edges]{forest}
\usepackage{microtype}      
\usepackage{placeins} 

\def\tsc#1{\csdef{#1}{\textsc{\lowercase{#1}}\xspace}}
\tsc{WGM}
\tsc{QE}
\tsc{EP}
\tsc{PMS}
\tsc{BEC}
\tsc{DE}

\begin{document}
\begin{sloppypar}
	\let\WriteBookmarks\relax
	\def\floatpagepagefraction{1}
	\def\textpagefraction{.001}
	\let\printorcid\relax
	\shorttitle{}
	\shortauthors{S. Zhang et~al.} 

	\title [mode = title]{Affective Computing in the Era of Large Language Models: A Survey from the NLP Perspective}



    \author[1]{\textcolor[RGB]{0,0,1}{Yiqun Zhang}}
	\fnmark[1]

    \author[1]{\textcolor[RGB]{0,0,1}{Xiaocui Yang}}
	\fnmark[1]
    
    \author[1]{\textcolor[RGB]{0,0,1}{Xingle Xu}}
    \fnmark[2]

    \author[1]{\textcolor[RGB]{0,0,1}{Zeran Gao}}
    \fnmark[2]

    \author[1]{\textcolor[RGB]{0,0,1}{Yijie Huang}}

    \author[1]{\textcolor[RGB]{0,0,1}{Shiyi Mu}}

    \author[1]{\textcolor[RGB]{0,0,1}{Shi Feng}}
    \cormark[1]
    \ead{fengshi@cse.neu.edu.cn}

     \author[1]{\textcolor[RGB]{0,0,1}{Daling Wang}}

     \author[1]{\textcolor[RGB]{0,0,1}{Yifei Zhang}}

     \author[2]{\textcolor[RGB]{0,0,1}{Kaisong Song}}

     \author[1]{\textcolor[RGB]{0,0,1}{Ge Yu}}

    \cortext[cor1]{Corresponding author.}
    \fntext[eq1]{These authors contributed equally to this work: Yiqun Zhang and Xiaocui Yang.}
    \fntext[eq2]{These authors contributed equally to this work: Xingle Xu and Zeran Gao.}
    
    \address[1]{Northeastern University, China}
    \address[2]{Alibaba Group, Hangzhou, China}






	\begin{abstract}
		Affective Computing (AC) integrates computer science, psychology, and cognitive science to enable machines to recognize, interpret, and simulate human emotions across domains such as social media, finance, healthcare, and education. AC commonly centers on two task families: Affective Understanding (AU) and Affective Generation (AG). While fine-tuned pre-trained language models (PLMs) have achieved solid AU performance, they often generalize poorly across tasks and remain limited for AG, especially in producing diverse, emotionally appropriate responses. The advent of Large Language Models (LLMs) (e.g., ChatGPT and LLaMA) has catalyzed a paradigm shift by offering in-context learning, broader world knowledge, and stronger sequence generation. This survey presents an NLP-oriented overview of AC in the LLM era. We (i) consolidate traditional AC tasks and preliminary LLM-based studies; (ii) review adaptation techniques that improve AU/AG, including Instruction Tuning (full and parameter-efficient methods such as LoRA, P-/Prompt-Tuning), Prompt Engineering (zero/few-shot, chain-of-thought, agent-based prompting), and Reinforcement Learning. For the latter, we summarize RL from human preferences (RLHF), verifiable/programmatic rewards (RLVR), and AI feedback (RLAIF), which provide preference- or rule-grounded optimization signals that can help steer AU/AG toward empathy, safety, and planning, achieving finer-grained or multi-objective control. To assess progress, we compile benchmarks and evaluation practices for both AU and AG. We also discuss open challenges—from ethics, data quality, and safety to robust evaluation and resource efficiency—and outline research directions. We hope this survey clarifies the landscape and offers practical guidance for building affect-aware, reliable, and responsible LLM systems.

	\end{abstract}


		
	\begin{keywords}
         Affective Computing \sep 
         Instruction Tuning \sep
         Prompt Engineering \sep
         Reinforcement Learning \sep
         Benchmarks
	\end{keywords}

	\maketitle

	\input{_1_intro}

        \input{_layout}

        \input{_2_tasks}
        \FloatBarrier
        
        \input{_3_preliminary}

        \input{_4_tuning}

        \input{_5_prompt}

        \input{_6_reinforcement}

        \input{_7_benchmark}

        \input{_8_future}
        
        \section{Conclusion}
        Affective Computing (AC) has emerged as a crucial research direction in artificial intelligence. With the rise of Large Language Models (LLMs), significant progress has been made in understanding and generating emotions. This paper provides a comprehensive review of LLM applications in AC, exploring the roles of techniques such as instruction tuning, prompt engineering, and reinforcement learning in affective understanding and generation tasks. We further conclude benchmarks for multiple large language models in AC to offer in-depth assistance to researchers and practitioners in related fields.
        Research indicates that LLMs, leveraging their powerful contextual learning and sequence generation capabilities, have demonstrated excellent affective understanding and generation performance.  However, LLMs still face numerous challenges in multilingual and multicultural sentiment analysis, real-time affective computing, and emotion-related tasks in specific vertical domains. We discuss challenges and future research directions related to AC, and aim to provide insights into this rapidly advancing field.
        
        \section*{Acknowledgments}
        Thanks to all co-authors for their hard work. 
        The work is supported by the National Natural Science Foundation of China (No.62272092, No.62172086, and No.62106039), and the Fundamental Research Funds for the Central Universities under Grants (N25XQD004, N25ZLL045).
        
	\bibliographystyle{model5-names}

	\bibliography{references}




\end{sloppypar}
\end{document}

%% file: _1_intro.tex
\section{Introduction}
\label{section:intro}
Affective Computing (AC), encompassing the study and development of systems capable of recognizing, interpreting, processing, and simulating human affects through natural language and other modalities, seeks to bridge the gap between human emotions and machine understanding~\cite{picard2000affective, tao2005affective, banafaWhatAffectiveComputing2016}.
The AC task can be broadly categorized into two mainstream tasks: the \textbf{Affective Understanding (AU)} task~\cite{zhaoChatGPTEquippedEmotional2023} and the \textbf{Affective Generation (AG)} task~\cite{nie2022review, 2024AffectiveComputingReview}. Affective Understanding focuses on recognizing and interpreting human emotions, covering Sentiment Analysis~\cite{medhat2014sentiment, taboada2016sentiment}, Sarcasm Detection~\cite{joshi2017automatic, verma2021techniques}, and so on.
Affective Generation involves creating responses or content that can elicit specific emotional responses in humans to generate expressions with emotionally nuanced, including Emotional-aware Dialogue Generation~\cite{rashkin2019empathetic, liu2021emotional}, Creative Content Generation, etc.
With the emergence of Large Language Models (LLMs)~\cite{openai2023gpt, touvron2023llama, zeng2023glm130b}, the Affective Computing task is experiencing a paradigm shift to explore and model human emotions in unprecedented ways.

\begin{figure*}
    \centering
    \includegraphics[width=\textwidth]{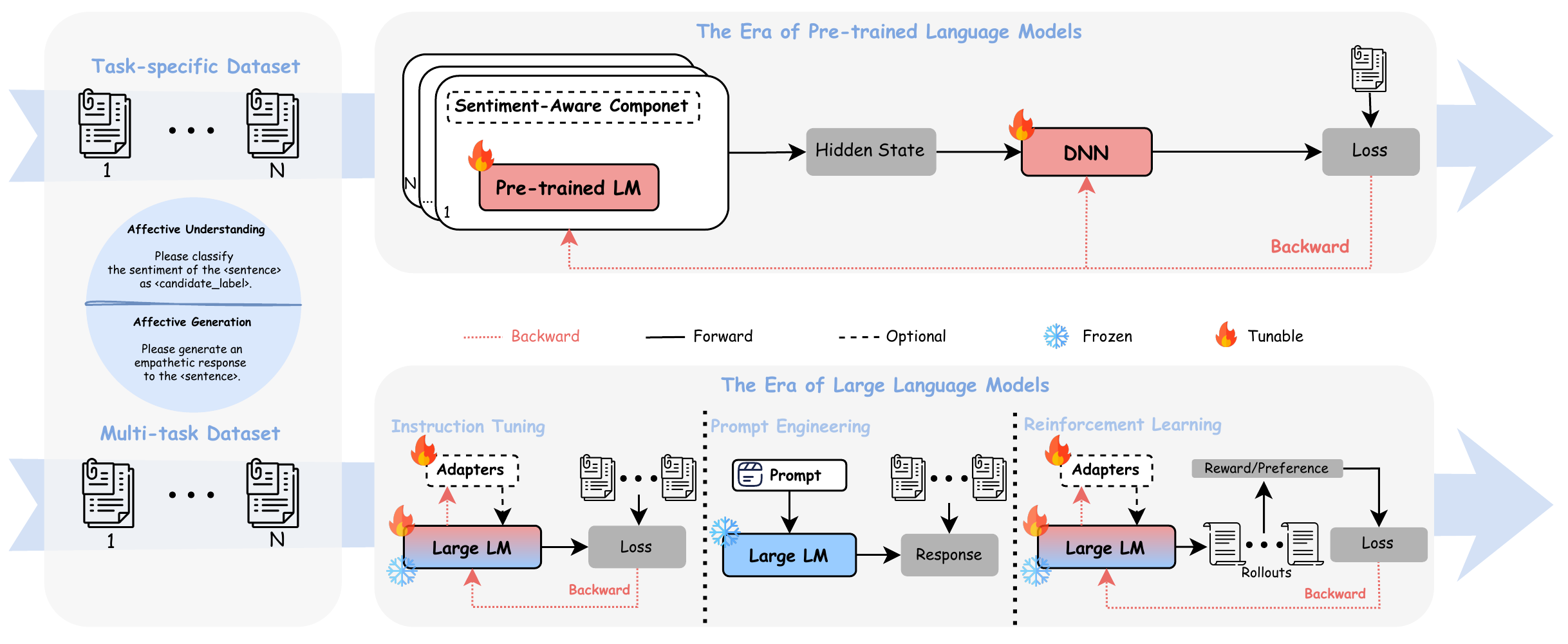}
    \caption{Comparison of approaches to AC tasks across the PLMs and LLMs eras. 
In the PLMs era, task-specific datasets fine-tune PLMs with sentiment-aware modules and downstream DNNs~\cite{bengio2009learning} (e.g., MLP~\cite{rosenblatt1958perceptron}, LSTM~\cite{hochreiter1997long}), typically requiring a separate model per task. 
In the LLMs era, a single (mostly frozen) LLM handles multiple tasks via (i) instruction tuning with adapters, (ii) prompt engineering, and (iii) reinforcement learning that samples rollouts and uses reward/preference signals to optimize a loss and update the adapters.}
    \label{fig: plm vs llm}
\end{figure*}

Prior to the advent of LLMs, the dominant paradigm of Affective Computing involves fine-tuning Pre-trained Language Models (PLMs)~\cite{2023willAffectiveComputing}, such as BERT~\cite{devlinBERTPretrainingDeep2019} and RoBERTa~\cite{liu2019roberta}, on labelled datasets for specific tasks to improve performance. 
While the above training paradigm has achieved numerous successes in the tasks related to Affective Understanding, it still is inherently limited by the quality and scale of manually 
annotated data and struggles to generalize to new domains~\cite{2022biasOfPLM}. 
The rise of LLMs research, as demonstrated in Figure \ref{fig: plm vs llm}, profoundly transforms the research paradigm of AC, expanding task variety and surpassing the constraints of previous studies, which often focus on particular datasets and single-task the performance improvements~\cite{app13074550, krugmannSentimentAnalysisAge2024}. This evolution is accompanied by developing novel approaches that exploit the vast knowledge of LLMs~\cite{wei2022emergent} to simultaneously boost the performance of various tasks related to Affective Computing.

The auto-regressive nature and massive pre-training of LLMs give them an advantage in processing dialogue inputs and generating coherent responses~\cite{radford2018improving, brown2020language}. 
Various techniques are proposed to fully harness and enhance the capabilities of LLMs, including \textbf{Instruction Tuning}, \textbf{Prompt Engineering} and \textbf{Reinforcement Learning}. 
Specifically, \textbf{Instruction Tuning}~\cite{chung2024scaling, zhang2024instruction} involves fine-tuning a pre-trained language model on the dataset where tasks are explicitly defined through instructions. For example, Flan-T5~\cite{chung2024scaling} designs specific instructions for different tasks, e.g., Sentiment Analysis, Commonsense Reasoning, Question Answering, etc.  Instruction Tuning is a form of targeted training that aims to make the model more adept at following complex and varied instructions across different domains.
\textbf{Prompt Engineering}~\cite{zhou2023largelanguagemodelshumanlevel, white2023prompt} is more about crafting the input given to the model to elicit the desired output without modifying the model itself. 
In-context Learning can further mine a model to learn and apply knowledge based on the instances provided directly within the prompt, without any prior explicit training on these specific tasks~\cite{xie2022explanation}. 
\textbf{Reinforcement Learning} optimizes the model against preference-based or verifiable rewards to improve helpfulness, safety, and style. A common pipeline, \emph{Reinforcement Learning from Human Feedback (RLHF)}, first trains a reward model from human comparisons and then updates the policy with online RL—typically \emph{Proximal Policy Optimization (PPO)}~\cite{lambert2025reinforcement}. Complementary directions include \emph{Reinforcement Learning with Verifiable Rewards (RLVR)}~\cite{lambert2024tulu}, which uses rule-checkers or pass/fail signals as rewards, and methods that reduce or avoid online RL, such as \emph{Direct Preference Optimization (DPO)}~\cite{rafailov2023dpo}. \emph{Reinforcement Learning from AI Feedback (RLAIF)} distills feedback from capable AI evaluators or rule sets into the preference signal (e.g.,~\cite{wen2025reinforcement}). For AC tasks, these approaches allow alignment directly toward affect-aware objectives, covering politeness, empathy, and non-toxic style.

Although LLMs offer new opportunities and show initial success in various tasks, a comprehensive review that consolidates and summarizes the recent advancements in AC with LLMs from an NLP perspective still needs to be present.
Existing surveys either focus narrowly on specific tasks like sentiment analysis or a limited set of classic affective computing tasks~\cite{app13074550, wakeBiasEmotionRecognition2023, cuiSurveySentimentAnalysis2023, zhang2023sentiment, zhu2024modelarenacrosslingualsentiment}, or they concentrate on traditional, non-LLM-based methods~\cite{2023willAffectiveComputing, afzal2023comprehensive}.
To mitigate the gap, it is essential to provide a comprehensive overview  (Figure~\ref{fig:overview}) of how LLMs are being applied across the diverse spectrum of tasks related to AC, along with the unique opportunities and challenges they bring.
Accordingly, we investigate the capabilities and limitations of LLMs in AC tasks, covering both 
Affective Understanding and Generation.
While our focus is primarily on NLP, we also include text-centric multimodal affective computing tasks, recognizing the growing importance of integrating multiple modalities in AC.
We begin by summarizing early studies that laid the groundwork for understanding the capabilities and limitations of LLMs in AC. Building on this, we explore three critical technologies for adapting LLMs to AC tasks: instruction tuning (\S\ref{sec:Instruction tuning}), prompt engineering (\S\ref{section:prompt_engineering}) and reinforcement learning (\S\ref{section:rl}). While tuning grounds the model in task formats and prompting controls behavior at inference, RL supplies preference- or programmatically verifiable objectives that close the optimization loop.
To deliver a comprehensive assessment, we further summarize benchmarks and evaluation methods (\S\ref{section:benchmark}) for systematically evaluating LLMs' performance in AC, including the innovative use of LLMs as evaluation tools. 
Finally, we enumerate open challenges and outline promising future research directions (\S\ref{sec:future}), offering insights into this rapidly advancing field's continuous evolution and potential.

%% file: _layout.tex
\definecolor{paired-light-blue}{RGB}{198, 219, 239}
\definecolor{paired-dark-blue}{RGB}{49, 130, 188}
\definecolor{paired-light-orange}{RGB}{251, 208, 162}
\definecolor{paired-dark-orange}{RGB}{230, 85, 12}
\definecolor{paired-light-green}{RGB}{199, 233, 193}
\definecolor{paired-dark-green}{RGB}{49, 163, 83}
\definecolor{paired-light-purple}{RGB}{218, 218, 235}
\definecolor{paired-dark-purple}{RGB}{117, 107, 176}
\definecolor{paired-light-gray}{RGB}{217, 217, 217}
\definecolor{paired-dark-gray}{RGB}{99, 99, 99}
\definecolor{paired-light-pink}{RGB}{222, 158, 214}
\definecolor{paired-dark-pink}{RGB}{123, 65, 115}
\definecolor{paired-light-red}{RGB}{231, 150, 156}
\definecolor{paired-dark-red}{RGB}{131, 60, 56}
\definecolor{paired-light-yellow}{RGB}{231, 204, 149}
\definecolor{paired-dark-yellow}{RGB}{141, 109, 49}
\definecolor{light-green}{RGB}{118, 207, 180}
\definecolor{raspberry}{RGB}{228, 24, 99}

\tikzset{%
    root/.style =          {align=center,text width=3cm,rounded corners=3pt, line width=0.5mm, fill=paired-light-gray!50,draw=paired-dark-gray!90},
    preliminary/.style =  {align=center,text width=4cm,rounded corners=3pt, fill=paired-light-blue!50,draw=paired-dark-blue!80,line width=0.4mm},
    sft/.style = {align=center,text width=4cm,rounded corners=3pt, fill=paired-light-orange!50,draw=paired-dark-orange!80,line width=0.4mm},
    prompt/.style = {align=center,text width=4cm,rounded corners=3pt, fill=paired-light-green!50,draw=paired-dark-green!80, line width=0.4mm},
    rl/.style = {align=center,text width=4cm,rounded corners=3pt, fill=paired-light-red!35,draw=paired-light-red!90, line width=0.4mm},
    benchmark/.style = {align=center,text width=4cm,rounded corners=3pt, fill=paired-light-purple!35,draw=paired-dark-purple!90, line width=0.4mm},
    subsection/.style =    {align=center,text width=3.5cm,rounded corners=3pt}, 
}

\begin{figure*}[!htb]
    \centering
    \resizebox{1\textwidth}{!}{
    \begin{forest}
        for tree={
            forked edges,
            grow'=0,
            draw,
            rounded corners,
            node options={align=center},
            text width=4cm,
            s sep=6pt,
            calign=child edge,
            calign child=(n_children()+1)/2,
            l sep=12pt,
        },
        [Affective Computing, root,
            [Preliminary Study (\S\ref{section:Preliminary Study}), preliminary
                [Affective Understanding (\S\ref{section1:au}), preliminary 
                    [Tan et al.~\cite{tan2023survey};
                    Rodriguez et al.~\cite{rodriguez2023review};
                    Das et al.~\cite{das2023multimodal};
                    Krugmann et al.~\cite{krugmannSentimentAnalysisAge2024};
                    Zhang et al.~\cite{zhangSentimentAnalysisEra2023};
                    Belal et al.~\cite{belalLeveragingChatGPTText2023};
                    Lossio et al.~\cite{lossio-venturaComparisonChatGPTFineTuned2024};
                    Alexander et al.~\cite{alexandersenManVsMachine2023};
                    Amin et al.~\cite{amin2023wide};
                    Wang et al.~\cite{wangChatGPTGoodSentiment2024};
                    Wu et al.~\cite{wuEnhancingLargeLanguage2024};
                    Zhang et al.~\cite{zhangRevisitingSentimentAnalysis2023};
                    M.M.Amin et al.~\cite{2023willAffectiveComputing},
                    preliminary, text width=12cm
                    ] 
                ]
                [Affective Generation (\S\ref{section1:ag}), preliminary
                    [
                    Zhao et al.~\cite{zhaoChatGPTEquippedEmotional2023};
                    Welivita et al.~\cite{welivitaChatGPTMoreEmpathetic2024};
                    Elyoseph et al.~\cite{elyosephChatGPTOutperformsHumans2023};
                    SECEU~\cite{wangEmotionalIntelligenceLarge2023};
                    EQ-Bench~\cite{paechEQBenchEmotionalIntelligence2024};
                    SOUL~\cite{dengSOULSentimentOpinion2023},
                    preliminary, text width=12cm
                    ] 
                ]
            ]
            [Instruction Tuning (\S\ref{sec:Instruction tuning}), sft
                [Parameter-Efficient Fine-Tuning, sft 
                    [
                    EMO-LLaMA~\cite{xing2024emo};
                    Emotion-LLaMA~\cite{cheng2024emotion};
                    InstructERC~\cite{lei2024instructerc};
                    CKERC~\cite{fu2024ckerc};
                    DialogueLLM~\cite{zhang2023dialoguellm};
                    Iain J. et. al~\cite{cruickshank2024promptingfinetuningopensourcedlarge};
                    STOEI~\cite{jia2024llm};
                    EKTC~\cite{cao-etal-2025-tool};
                    Wang et al.~\cite{wang2023enhancing};
                    PEGS~\cite{zhang2024stickerconv};
                    GSA~\cite{hou-etal-2024-progressive};
                    Penf et al.~\cite{peng2024customising};
                    Ding et al.~\cite{ding-etal-2024-boosting};
                    M2SE~\cite{li2024m2se};
                    MoEI~\cite{zhao2024both};
                    Zhao et al.~\cite{zhao2024both};
                    PICA~\cite{zhang2023PICA};
                    Zheng et al.~\cite{zheng2023building};
                    BLSP-Emo~\cite{wang2024blsp};
                    EMOLLM~\cite{liu2024emollms};
                    FAME-Net~\cite{wang2024generative};
                    PerceptiveAgent~\cite{yan2024talk},
                    sft, text width=12cm] 
                ]
                [Full-Parameter Fine-Tuning, sft 
                    [Colin Raffel et al.~\cite{2020t5};
                    Boyu Zhang et al.~\cite{zhang2023enhancingfinancialsentimentanalysis};
                    USA~\cite{gan2023usa};
                    Wisdom~\cite{wang2024wisdom};
                    Omni-Emotion~\cite{yang2025omni};
                    InstructABSA~\cite{scaria2023instructabsa};
                    ITSCL~\cite{zhang-etal-2024-instruction};
                    Siddharth Varia et al.~\cite{varia2023instruction};
                    UniSA~\cite{li2023unisa};
                    NUS-Emo~\cite{luo2024nus};
                    ObG~\cite{wang2024observe};
                    SoulChat~\cite{chen2023soulchat};
                    EMOVA~\cite{chen2024emova};
                    Emollms~\cite{liu2024emollms};
                    MODA~\cite{zhang2025modamodularduplexattention};
                    Emoada~\cite{dong2024emoada};
                    Lim Dongjun et al.~\cite{lim-cheong-2024-integrating},
                    sft, text width=12cm] 
                ]
            ]
            [Prompt Engineering (\S\ref{section:prompt_engineering}), prompt
                [Zero-shot (\S\ref{section:zero-shot for au} \& \ref{section:zero-shot for ag}), prompt
                [
                Amin et al.~\cite{amin2023wide};
                Ouyang et al.~\cite{ouyang2024stability};
                M.A.Al Asad et al.~\cite{al2023sentiment};
                W.Stigall et al.~\cite{10.1145/3603287.3651183};
                Hong et al.~\cite{hong2025aer};
                Zhou et al.~\cite{zhou2023prompt};
                Bai et al.~\cite{bai2024compound};
                Cabello et al.~\cite{cabello2024simple};
                Wang et al.~\cite{wang2023enhance}
                Niu et al.~\cite{niu2024text};
                Plaza et al.~\cite{plaza2024angry};
                Li et al.~\cite{li2023large};
                Shen et al.~\cite{shen2024heart};
                Wu et al.~\cite{wu2024beyond};
                Santoso et al.~\cite{santoso2024large};
                Liang et al.~\cite{liang2024aligncap};
                Wang et al.~\cite{wang2024reasoning}
                Lee et al.~\cite{lee2023investigating};
                Yang et al.~\cite{yang2024enhancing};
                TCG~\cite{bhaskar2022prompted};
                Y.M.Resendiz~\cite{resendiz2023emotion};
                PromptMind~\cite{su2023prompt},
                prompt, text width=12cm]]
                [Few-shot (\S\ref{section:fewshot for au} \& \ref{section:fewshot for ag}), prompt
                [
                Zhou et al.~\cite{zhou2024comprehensive};
                Zhang et al.~\cite{zhang2024instruction};
                Yang et al.~\cite{yang2024empirical};
                SCRAP~\cite{kim2024self};
                Wisdom~\cite{yang2024empirical};
                LVLMS~\cite{liu2024visual, ye2024mplug};
                SEGA~\cite{chen2024depression};
                Sibyl~\cite{Wang2023SibylSE};
                Chen et al.~\cite{chen2023controllable};
                Kang et al.~\cite{kang2024can};
                LLaMA2~\cite{touvron2023llama},
                prompt, text width=12cm]]
                [Chain-of-Thought (\S\ref{section:cot for au} \& \ref{section:cot for ag}), prompt
                [
                Lee et al.~\cite{lee2024analyzing};
                THOR~\cite{fei2023reasoning};
                Lyu et al.~\cite{lyu2024llms};
                CoE~\cite{lee2023chain};
                Manzoor et al.~\cite{manzoor2024can};
                ECR-Chain~\cite{huang2024ecr};
                RP-CoT~\cite{wang2023enhance};
                DECC~\cite{wu2024enhancing};
                EMO~\cite{baumann2024evolutionary};
                Deng~\cite{deng2023llms};
                CFEG~\cite{chen2024cause};
                Cue-CoT~\cite{hongru2023cue};
                EDIT~\cite{wu2023new};
                Aptness~\cite{hu2024aptness};
                ProCoT~\cite{deng2023prompting};
                ECoT~\cite{li2024enhancing};
                ESCoT~\cite{zhang2024escot};
                Li et al.~\cite{li2024sentiment};
                VEHICLE~\cite{shao2024cmdag},
                prompt, text width=12cm]]
                [Agent-based (\S\ref{section:agent for au} \& \ref{section:agent for ag}), prompt
                [
                Fei et al.~\cite{fei2023reasoning};
                Yang et al.~\cite{yang2024empirical};
                Sun et al.~\cite{sun2023sentiment};
                Wei et al.~\cite{wei2024mimicking};
                E.S.Van et al.~\cite{van2023empirical};
                PANAS~\cite{regan2024can};
                HAD~\cite{xing2024designing};
                Agent4SC~\cite{zhang2024stickerconv};
                COOPER~\cite{cheng2024cooper};
                PerceptiveAgent~\cite{yan2024talk};
                Emotion-aware agents~\cite{hu2022acoustically};
                Zhang et al.~\cite{zhang2024self},
                prompt, text width=12cm]]
            ]
            [Reinforcement Learning
            (\S\ref{section:rl}), rl
                [RLHF(\S\ref{section:rlhf for rl}), rl
                [
                Zhu et al.~\cite{zhu2023grafting};
                XU et al.~\cite{xu2025rlthf};
                AlignCap~\cite{liang2024aligncap};
                Chen et al.~\cite{chen2025empathyagent},
                rl, text width=12cm]
                ] 
                [RLVR(\S\ref{section:rlvr for rl}), rl
                [
                R1-Omni~\cite{zhao2025r1};
                SAGE (State Augmented GEneration)~\cite{zhang2025sage};
                 SAGE (Sentient Agent as a Judge)~\cite{zhang2025sentient};
                CRP~\cite{li2024helpful};
                EmpRL~\cite{ma2025empathy};
                ReflectDiffu~\cite{yuan2024reflectdiffu};
                ABLE~\cite{mishra2024able};
                GENTEEL-NEGOTIATOR~\cite{priya2025genteel};
                F$^2$RL~\cite{wang2024f2rl};
                Unnikrishnan et al.~\cite{unnikrishnan2024financial},
                rl,text width=12cm]               
                ]
                [RLAIF(\S\ref{section:rlaif for rl}), rl
                [
                Yoshida et al.~\cite{yoshida2025training}
                PPDPP~\cite{deng2023plug};
                DialogXpert~\cite{rakib2025dialogxpert};
                CSO~\cite{zhao2025chain};
                Ye et al.~\cite{ye2025generic};
                Decoupled ESC~\cite{zhang2025decoupledesc},
                rl,text width=12cm]               
                ]
            ]
            [Benchmark \& Evaluation
            (\S\ref{section:benchmark}), benchmark
                [Affective Understanding (\S\ref{section:au for benchmark}), benchmark
                [SOUL~\cite{dengSOULSentimentOpinion2023};
                GPT-4v with Emotion~\cite{Lian2023GPT4VWE};
                MERBench~\cite{Lian2024MERBenchAU};
                DFEW~\cite{jiang2020dfew};
                MC-EIU~\cite{Liu2024EmotionAI},
                benchmark, text width=12cm]
                ] 
                [Affective Generation (\S\ref{section:ag for benchmark}), benchmark
                [
                Cue-CoT~\cite{hongru2023cue};
                ESC-Eval~\cite{zhao2024esc};
                MEDIC~\cite{Zhu2023MEDICAM};
                EmotionBench~\cite{Huang2023EmotionallyNO};
                STICKERCONV~\cite{zhang2024stickerconv};
                SECEU~\cite{Wang2023EmotionalIO};
                EQ-Bench~\cite{paechEQBenchEmotionalIntelligence2024};
                EmoBench~\cite{Sabour2024EmoBenchET};
                EIBENCH~\cite{zhao2024both};
                EmoVIT~\cite{Xie2024EmoVIT};
                EMER~\cite{lian2023explainable};
                MER-UniBench~\cite{lian2025affectgpt},
                benchmark, text width=12cm]
                ]
                [Others (\S\ref{section:others for benchmark}), benchmark
                [
                ChatGPT2AC~\cite{aminWideEvaluationChatGPT2023};
                MM-INSTRUCTEVAL~\cite{Yang2024MMInstructEvalZE},
                benchmark, text width=12cm]
                ]
            ]
        ]
    \end{forest}
    }
    \caption{The overview of the paper.} 
    \label{fig:overview}
\end{figure*}

%% file: _2_tasks.tex
\section{Tasks} 
\label{section:Tasks}
Affective Computing (AC) comprises two mainstream task families: Affective Understanding (AU)~\cite{zhaoChatGPTEquippedEmotional2023} and Affective Generation (AG)~\cite{nie2022review, 2024AffectiveComputingReview}, as summarized in Table~\ref{tab: task overview}.

\begingroup

\begin{table*}[t]
\renewcommand\arraystretch{1.2}
\centering
  \caption{The traditional tasks of Affective Computing from an NLP perspective, with representative methods in the LLM era.}
  \resizebox{1.0\textwidth}{!}{
  
\begin{tabular}{c|c|c|c}
\toprule[1pt]
\textbf{AC} & \textbf{Tasks} & \textbf{Subtasks}    & \textbf{Examples}        \\ 
\midrule[1pt]
\multirow{13}{*}{\textbf{AU}} & \multirow{7}{*}{Sentiment Analysis (SA)}        & Polarity Classification (PC) &  USA~\cite{gan2023usa}, WisdoM~\cite{wang2024wisdom}, RP-CoT~\cite{wang2023enhance}, Sun et.al~\cite{sun2023sentiment}     \\ \cline{3-4} 
            &                & Emotional   Classification (EC)            &  Zhang et.al~\cite{zhang2023sentiment}, Stigall et.al~\cite{10.1145/3603287.3651183} \\ \cline{3-4} 
            &                & Aspect-Based Sentiment Analysis (ABSA) & InstructABSA~\cite{scaria2023instructabsa}, SCRAP~\cite{kim2024self} \\ \cline{3-4} 
            &                & Emotional Intensity Detection   (EID)      & Amin et.al~\cite{amin2023wide} \\ \cline{3-4} 
            &                & Implicit Sentiment Analysis   (ISA)        & THOR~\cite{fei2023reasoning} \\ \cline{3-4} 
            &                & Emotion Recognition in   Conversation (ERC) & UniMSE~\cite{hu2022unimse}, DialogueLLM~\cite{zhang2023dialoguellm}, CKERC~\cite{fu2024ckerc}, InstructERC~\cite{lei2024instructerc} \\ \cline{3-4} 
            &                & Emotion Cause Paired Extraction (ECPE)      &  DECC~\cite{wu2024enhancing}, ECR-Chain~\cite{huang2024ecr}\\ \cline{2-4} 
                     & \multirow{6}{*}{Subjective Text Analysis (STA)} & Suicide Tendency Detection (STD)        &     Zhang et.al~\cite{zhang2023sentiment}, Amin et.al~\cite{amin2023wide}  \\  \cline{3-4} 
            &                & Personality Assessment (PA)                &  Zhang et.al~\cite{zhang2023sentiment}, Amin et.al~\cite{amin2023wide} \\  \cline{3-4} 
            &                & Toxicity Detection (TD)                    &  Zhang et.al~\cite{zhang2023sentiment}, Amin et.al~\cite{amin2023wide} \\  \cline{3-4} 
            &                & Sarcasm Detection (SD)                     &  Hoffmann et.al~\cite{hoffmann2022training}, Zhang et.al~\cite{zhang2023sentiment}, Amin et.al~\cite{amin2023wide} \\  \cline{3-4} 
            &                & Well-Being Assessment (WBA)                &  Zhang et.al~\cite{zhang2023sentiment}, Amin et.al~\cite{amin2023wide}  \\  \cline{3-4} 
            &                & Engagement   Measurement (EM)              & Zhang et.al~\cite{zhang2023sentiment}, Amin et.al~\cite{amin2023wide}  \\  
\midrule[1pt]
\multirow{3}{*}{\textbf{AG}}  & \multirow{2}{*}{Emotional Dialogue (ED)}     & Empathetic Response Generation (ERG) & Sibyl~\cite{wang2023enhancing}, PromptMind~\cite{yang2024enhancing}, PEGS~\cite{zhang2024stickerconv}, Chain of Empathy~\cite{lee2023chain}, ECoT~\cite{Li2024EnhancingEG} \\ \cline{3-4} 
            &                & Emotional   Support Conversation (ESC)     & SoulChat~\cite{chen2023soulchat}, EmoLLM~\cite{EmoLLM}, PICA~\cite{zhang2023PICA}, Cue-CoT~\cite{hongru2023cue}  \\ \cline{2-4} 
                     & Review   Summarization (RS)                     & Opinion Summarization (OS)        &   TCG~\cite{bhaskar2022prompted}, LASS~\cite{li2023large} \\ 
\bottomrule[1pt]
\end{tabular}

}
  \label{tab: task overview}%
\end{table*}%
\endgroup

\subsection{Affective Understanding}
\subsubsection{Sentiment Analysis}
AU focuses on recognizing and interpreting human affect, with Sentiment Analysis (SA) as the core task. The basic form is Polarity Classification (PC)~\cite{pang2002thumbs}, which labels a text as negative, positive, or neutral. Emotional Classification (EC)~\cite{li2017dailydialog, demszky2020goemotions} extends this to finer‑grained emotion categories.
Aspect-Based Sentiment Analysis (ABSA)~\cite{schouten2015survey, 2023absa} targets aspect-level opinions. Typical subtasks include Aspect Term Extraction (ATE)~\cite{toh2014dlirec, li2018aspect} and Aspect-oriented Sentiment Classification (ASC)~\cite{lau2014social, lau2018parallel}; Joint Aspect–Sentiment Analysis (JASA)~\cite{zhuang2020joint} models both simultaneously, and ACOS~\cite{cai2021aspect} further incorporates category and opinion dimensions.
Beyond label granularity and aspects, Emotional Intensity Detection (EID)~\cite{bonnet2015role, alonso2015new} estimates strength, Implicit Sentiment Analysis (ISA)~\cite{tubishat2018implicit, wei2020bilstm} captures sentiment not explicitly expressed, and Multimodal Sentiment Analysis (MSA)~\cite{das2023multimodal, jung2022multimodalemotion} fuses textual, visual, and acoustic cues.

Conversation-level understanding has become central with advances in human–computer interaction~\cite{chen2017survey}. Emotion Recognition in Conversation (ERC)~\cite{li2017dailydialog, poria2019emotion, poria2018meld} predicts utterance-level emotions while tracking dynamics in context. Emotion and Intent Joint Understanding in Multimodal Conversation (MC-EIU) jointly infers current-utterance emotion and intent~\cite{Liu2024EmotionAI}. Emotion Cause Pair Extraction (ECPE)~\cite{xia2019emotion} extracts emotions together with their causes.
See Table~\ref{tab: task overview} for the complete taxonomy and representative methods.

\subsubsection{Subjective Text Analysis}
Subjective Text Analysis (STA) probes user traits, risks, and attitudes expressed in text, including Personality Assessment~\cite{ponce2016chalearn}, Suicide Tendency Detection~\cite{10.1007/978-981-16-9705-0_26}, Toxicity Detection\footnote{https://kaggle.com/competitions/jigsaw-toxic-comment-classification-challenge}, Sarcasm Detection~\cite{misra2023sarcasm}, Well-being Assessment~\cite{rastogi2022stress}, and Engagement Measurement\footnote{https://www.kaggle.com/datasets/thedevastator/social-media-interactions-on-tedtalks-dataset}. 
Compared with classic SA, these tasks are relatively under-explored and under-resourced.

\subsection{Affective Generation}
\subsubsection{Emotional Dialogue}
Affective Generation (AG) aims to produce emotionally appropriate, empathetic, or supportive content. In Emotional Dialogue, Empathetic Response Generation (ERG)~\cite{rashkin2019empathetic} generates context-aware empathetic replies, while Emotional Support Conversation (ESC)~\cite{liu2021emotional} targets goal-directed support. Multimodal variants (e.g., MERG) leverage audio/vision with text to enhance empathy and naturalness~\cite{jung2022multimodalemotion, zhang2024stickerconv}.

\subsection{Review Summarization}
Review Summarization (RS) is a broader task where Opinion Summarization (OS)~\cite{kim2011comprehensive, moussa2018survey} condenses many user opinions on a topic—often from social media or UGC—into concise summaries that surface overall sentiment and key points. Challenges arise from noisy, unstructured text and figurative or diverse language use (e.g., sarcasm).

It is worth noting that research on AG lagged behind AU in the pre-LLM era due to limitations in model capabilities and technology. 
However, with their exceptional sequence generation abilities, the advent of LLMs has ushered in new possibilities for affect generation, making it a focal point of research in the LLM era~\cite{park2023generative}.

%% file: _3_preliminary.tex
\section{Preliminary Study}
\label{section:Preliminary Study}
Large Language Models (LLMs), with their powerful natural language processing, are rapidly transforming the field of Affective Computing. This section explores the emerging applications of LLMs in crucial areas of Affective Computing: Affective Understanding and Affective Generation. We review the current state of research, highlighting the strengths and limitations of LLMs in these tasks and suggest promising directions for future exploration.

\subsection{Affective Understanding}
\label{section1:au}

\subsubsection{Sentiment Analysis}
Sentiment Analysis (SA), a classic task in the Affective Computing domain, focus on the automatic identification and categorization of emotions and sentiments expressed through contents from users~\cite{tan2023survey, rodriguez2023review, das2023multimodal}.

To evaluate the performance of LLMs in SA tasks, researchers conduct assessments on multiple datasets and discover that the performance of LLMs is closely related to the complexity of the SA tasks they face.
For more straightforward sentiment analysis tasks, such as Polarity Classification (PC), closed-source large language models like GPT-3.5-turbo demonstrate robust zero-shot learning capabilities and can deliver satisfactory results~\cite{krugmannSentimentAnalysisAge2024,zhangSentimentAnalysisEra2023,belalLeveragingChatGPTText2023, lossio-venturaComparisonChatGPTFineTuned2024, alexandersenManVsMachine2023}.
However, when confronted with complex SA tasks requiring deep semantic understanding or structured emotional information, such as Emotional Classification (EC), Aspected-Based Sentiment Analysis (ABSA), Emotional Recognition in Conversation (ERC), etc., which involves identifying multiple sentiments corresponding to different aspects of a post, or Emotion Cause Analysis, the performance of LLMs still lags behind that of pre-trained language models (PLMs) fine-tuned on specific datasets~\cite{zhangSentimentAnalysisEra2023, aminWideEvaluationChatGPT2023, wangChatGPTGoodSentiment2024, wuEnhancingLargeLanguage2024}. This discrepancy may stem from LLMs lacking training data tailored to specific tasks and having insufficient capabilities to process complex semantic information.

Although the zero-shot learning capabilities of open-source LLMs are inferior to those of closed-source models, their superior few-shot learning abilities give them an advantage on datasets with limited training data or imbalanced distributions, outperforming PLMs in such scenarios. However, when data is abundant and label distribution is balanced, PLMs can still perform better due to their fine-tuning advantage on specific datasets~\cite{zhangRevisitingSentimentAnalysis2023}.
Notably, LLMs show superior generalization capabilities in SA tasks compared to PLMs. By simply altering the prompts, LLMs can adapt to different datasets and tasks without requiring model retraining~\cite{wangChatGPTGoodSentiment2024}.
However, the performance of LLMs is also affected by prompts and decoding parameters. For example, minor text attacks like synonym substitution can lead to fluctuations in their performance. Additionally, because of the auto-regressive decoding mechanism, parameters such as temperature and Top-P can significantly impact the effectiveness of LLMs in sentiment analysis tasks. 
Typically, a lower temperature value (below 0.3) and a lower Top-P value are necessary for achieving better results~\cite{2023willAffectiveComputing, aminPromptSensitivityChatGPT2024}.  
This suggests that although research shows carefully crafted prompts can greatly improve LLMs'performance, developing optimal prompts and setting effective hyper-parameters for specific tasks and contexts remains an unresolved research challenge.

\subsubsection{Subjective Text Analysis}

LLMs demonstrate considerable performance variations across various Subjective Text Analysis (STA) related tasks. For instance, they excel in tasks involving negative emotion recognition, such as well-being assessment and toxicity detection. The superior performance could be due to the focus on safety and value alignment during the Reinforcement Learning from Human Feedback (RLHF) stage of LLM training~\cite{zhengSecretsRLHFLarge2023, 2023willAffectiveComputing, aminWideEvaluationChatGPT2023}.
However, in tasks requiring the understanding of implicit emotional cues, such as engagement measurement, personality assessment, and sarcasm detection, LLMs exhibit a performance gap compared to PLMs fine-tuned on task-specific datasets. 
This discrepancy may stem from a need for more specialized training data during the supervised fine-tuning (SFT) phase of LLM development. 
Optimizing prompts for LLMs can enhance their performance on these specific tasks~\cite{2023willAffectiveComputing,aminWideEvaluationChatGPT2023}.
Furthermore, leveraging their robust capacity for capturing contextual information, LLMs consistently demonstrate superior performance in processing longer texts compared to shorter texts across various subjective text analysis tasks~\cite{aminWideEvaluationChatGPT2023, krugmannSentimentAnalysisAge2024}. 
LLMs perform slightly worse than SFT models tailored to specific tasks in complex sentiment computing areas like sarcasm detection and subtle sentiment analysis. However, the substantial computational and time resources needed to fine-tune these large language models present a significant hurdle. 
Consequently, developing LLMs with outperformance and minimal training for specific tasks is an important research priority.

\subsection{Affective Generation}
\label{section1:ag}
\subsubsection{Emotional Dialogue}
Empathetic Response Generation (ERG) and Emotional Support Conversation (ESC) are two prominent tasks in AG that attract significant attention, utilizing the EMPATHETICDIALOGUES~\cite{rashkin2019empathetic} and ESConv~\cite{liu2021emotional}, respectively.
Early research suggests that LLMs possess immense potential in emotionally-aware generation tasks. For example, GPT-3.5-turbo has shown competitive performance across multiple datasets~\cite{zhaoChatGPTEquippedEmotional2023}, while GPT-4 generates empathetic responses in various emotional scenarios. Human evaluations indicate that GPT-4's responses exceed human performance by an average of 10\% in empathy levels~\cite{welivitaChatGPTMoreEmpathetic2024}.
Unlike earlier models, LLMs produce longer and more diverse text~\cite{welivitaChatGPTMoreEmpathetic2024}, introducing new evaluation challenges. 
Traditional automatic evaluation methods based on overlap metrics need to adequately measure the performance of LLMs in emotional generation tasks, while manual evaluation is costly and burdensome to scale. 
Developing novel evaluation methods and establishing standardized frameworks to accurately measure LLM-generated affective responses' quality, empathy, and emotional intelligence is crucial for advancing research in this area.

\subsubsection{Emotional Intelligence}
Emotional Intelligence (EI)~\cite{dulewicz2000emotional, mayer2003measuring, neubauer2005models} is the ability to recognize, understand, manage, and use emotions effectively in various contexts.
EI is crucial for enhancing the naturalness and effectiveness of human-computer interaction and is a vital component in achieving Artificial General Intelligence (AGI). To explore whether LLMs possess EI, researchers undertake investigations from various perspectives. Current studies~\cite{elyosephChatGPTOutperformsHumans2023, ratican2023six} draw upon assessment methods from the field of human psychology, employing tools, like the Levels of Emotional Awareness Scale (LEAS)~\cite{lane1990levels}, to evaluate the ability of LLMs to perceive emotions. LLMs can generate relatively appropriate emotional responses, which has been evidenced.

However, using human-centric evaluation systems to measure the EI of LLMs has limitations, including the fact that the scales are designed with humans in mind, not explicitly specifically for LLMs and that the evaluations rely on human experts to conduct the assessments. Researchers have developed evaluation tools designed explicitly for LLMs to address these issue, such as the Situational Evaluation of Complex Emotional Understanding (SECEU)~\cite{wangEmotionalIntelligenceLarge2023}.
Moreover, to improve the objectivity and efficiency of evaluations, EQ-bench has been proposed~\cite{paechEQBenchEmotionalIntelligence2024}. It removes the need for expert assessments and instead directly evaluates the ability of LLMs  to understand complex emotions and social interactions by predicting the intensity of emotional states in dialogue characters.
While LLMs have demonstrated remarkable performance on SECEU and EQBench, which focus more on the cognition and understanding of human emotional scenarios, there is still insufficient evidence to suggest that they possess emotional intelligence comparable to humans.
To address this gap, researchers propose a new task called Sentiment and Opinion Understanding of Language (SOUL), which aims to comprehensively evaluate the model's ability to understand complex emotions through two sub-tasks, including Review Comprehension and Justification Generation. However, experimental results reveal a substantial performance gap of up to 27\% between LLMs and human performance on the SOUL~\cite{dengSOULSentimentOpinion2023}. It highlights the considerable room for improvement in the capacity of LLMs to understand the nuances and complexities of human emotions.

%% file: _4_tuning.tex
\section{Instruction tuning}
\label{sec:Instruction tuning}

\begin{figure*}
	\centering
	\begin{subfigure}{0.25\textwidth}
		\centering
		\includegraphics[width=1.0\textwidth]{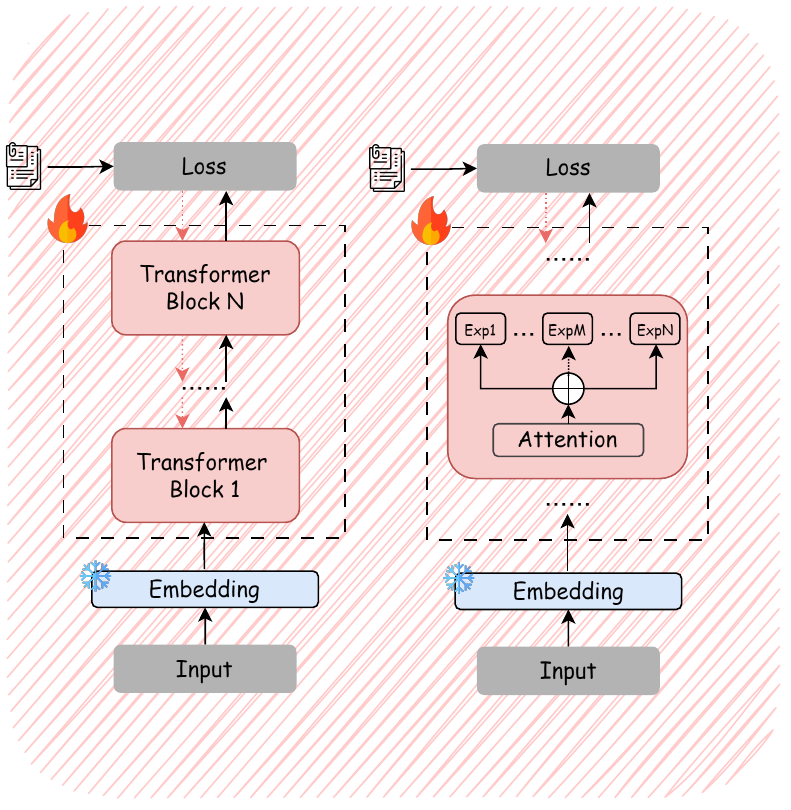}
		\caption{Full-Model Tuning}
		\label{subA}
	\end{subfigure}
	\centering
	\begin{subfigure}{0.30\textwidth}
		\centering
		\includegraphics[width=1.0\textwidth]{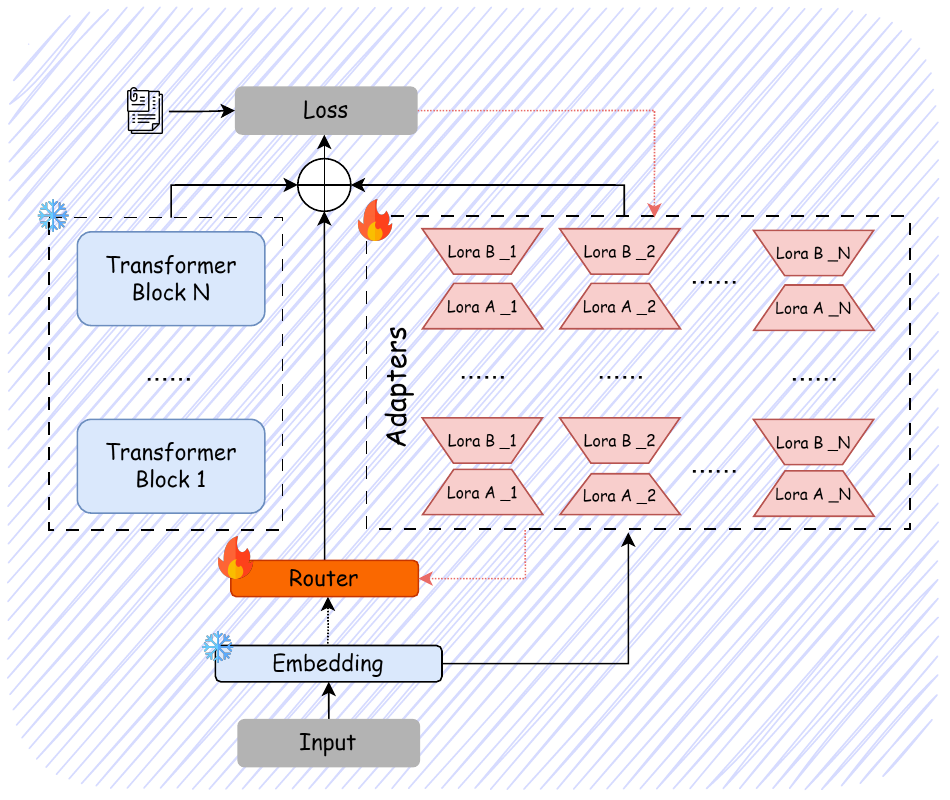}
		\caption{Multi-LoRA}
		\label{subB}
	\end{subfigure}
	\centering
	\begin{subfigure}{0.20\textwidth}
		\centering
		\includegraphics[width=1.0\textwidth]{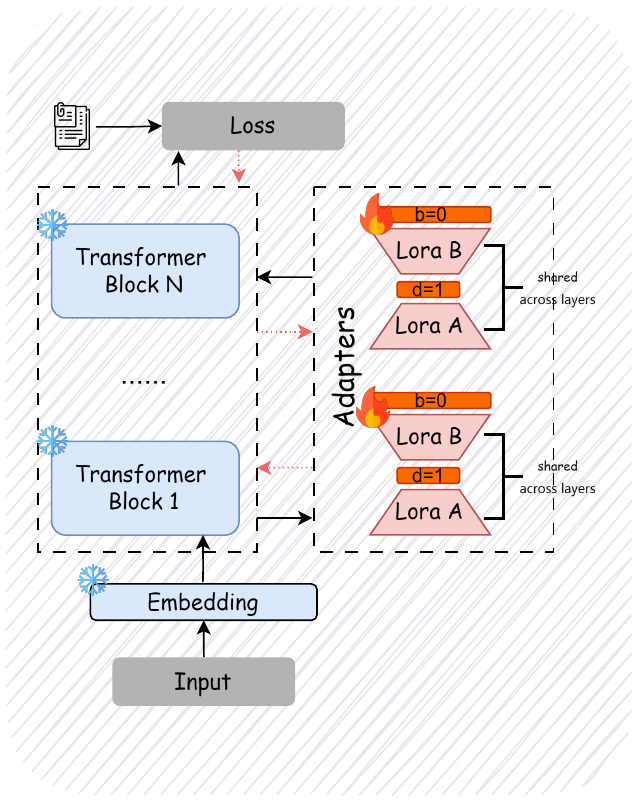}
		\caption{LoRA}
		\label{subC}
	\end{subfigure}
        \centering
	\begin{subfigure}{0.15\textwidth}
		\centering
		\includegraphics[width=1.0\textwidth]{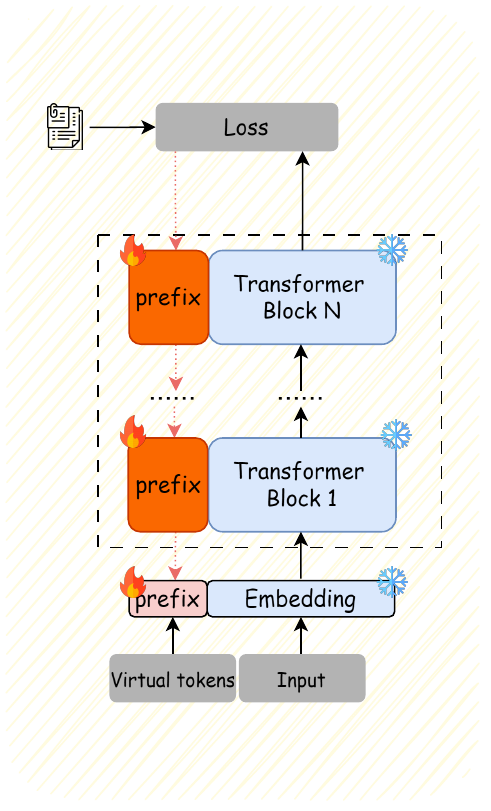}
		\caption{Prompt Tuning}
		\label{subD}
	\end{subfigure}
	\caption{Illustrations of the instruction tuning methods. Pink blocks denote trainable modules, blue blocks signify frozen modules, and orange blocks represent optionally added modules. Subfigure \ref{subA} depicts the FPFT method, the transformer blocks of two distinct model architectures are fine-tuned. Subfigure \ref{subB} introduces the multi-LoRA method. In this approach, only the added multi-LoRA models are trained. This method allows for the optional incorporation of router modules to manage the weights of each LoRA module. Subfigure \ref{subC} demonstrates the LoRA method, The traditional LoRA method involves the addition of only two low-rank matrix modules, LoRA\_A and LoRA\_B. VeRA \cite{kopiczko2024veravectorbasedrandommatrix} allows the addition of two trainable vectors, d and b, augmenting the LoRA layer. This approach also shares LoRA\_A and LoRA\_B's parameters to every LoRA layer. Moreover, when the rank of LoRA in each layer varies, this method can transition into adaLoRA \cite{zhang2023adaloraadaptivebudgetallocation}. Subfigure \ref{subD} illustrates the soft prompt tuning, where a trainable prefix is appended to the model's embedding layer, known as p-tuning. Furthermore, scientific prefixes can be optionally added to each layer of the model, with only the added prefixes being trained, a technique referred to as prefix-tuning.}
	\label{fig: tuning}
\end{figure*}

LLMs show excellent performance on different tasks, such as Question Answering, Sequence Generation, and so on \cite{zhao2023survey}. 
Instruction Tuning \cite{zhang2023instruction} is introduced to adapt LLMs to specific tasks, thereby maintaining the benefits of the original model while improving the performance of LLMs on specialized downstream tasks.
Due to the substantial size of LLMs, which often exceed ten billion parameters, Full Parameter Fine-Tuning (FPFT) of all parameters of the complete model could be more practical. To reduce the cost of model training, a series of Parameter-Efficient Fine-Tuning (PEFT) approaches are proposed, such as LoRA \cite{8474715}, Prefix Tuning \cite{li2021prefixtuning}, and P-Tuning \cite{liu-etal-2022-p}, as shown in Figure \ref{fig: tuning}.
LoRA reduces the number of parameters that need to be trained during the fine-tuning process by freezing all the original parameters and injecting a pair of low-rank decomposition matrices next to the original weights.
Prefix Tuning adds trainable "prefix" tokens related to specific tasks as an implicit learnable prompt to the input sequence of the LLM, inserting trainable free vectors into each layer of the Transformer. Applying PEFT in instruction tuning can effectively reduce the training costs in AC tasks.

\subsection{Affective Understanding} 

In addressing affective computing tasks, PLMs typically necessitate separate approaches for affective understanding and generation tasks. In contrast, LLMs can unify all affective computing tasks into a sequence generation format. 
However, directly employing LLMs for AU tasks does not offer significant advantages over fine-tuned PLMs.
Consequently, numerous recent studies focus on constructing task-specific instruction tuning templates tailored to the unique characteristics of various tasks. These studies apply instruction tuning to the models, aiming to enhance the performance of LLMs on specific tasks or within particular domains.

\subsubsection{Polarity and Emotional Classification} 
Polarity Classification (PC) and Emotional Classification (EC) tasks are frequently employed to assess a model's comprehension and perceptual abilities. 
Several studies pioneer the exploration of the effects of instruction tuning of LLMs on the SA task~\cite{raffel2020exploring,peng2024customising}. The T5~\cite{2020t5} series model, in particular, demonstrates an improved understanding of the sentiment content of text after instruction tuning.
Recent research \cite{peng2024customising} focuses on exploring the different PEFT methods, adopting the LoRA and P-tuning methods to fine-tuning ChatGLM2~\cite{glm2024chatglm}. Experimental results demonstrate that after instruction tuning, LLMs are greatly improved in eight datasets compared with the model without instruction tuning. However, the experimental results must conclusively determine which instruction tuning method is more effective in enhancing performance on PC. 

While these studies demonstrate LLMs' effectiveness in PC and EC tasks, it is equally important to understand their underlying sentiment processing mechanisms. Research shown that words with sentiment information can influence the overall sentiment of the text \cite{gan2023usa}.
Building on this understanding, LLMs leverage techniques of instruction tuning to grasp word polarity, enabling them to effectively perform methods PC and EC tasks. 
These studies primarily focus on text-based PC and EC tasks, it is essential to recognize that real-world user expressions are not limited to text alone. User-generated content often encompasses multiple modalities, including images and audio, which can provide additional context and nuance to sentiment analysis. 
WisdoM \cite{wang2024wisdom} leverages contextual world knowledge derived from MLLM to enhance multimodal sentiment analysis (MSA). Additionally, the authors introduce a training-free module called "contextual fusion", which can reduce noise in the contextual data and improve model classification accuracy. 
WisdoM significantly outperforms existing state-of-the-art methods on the MSED dataset, demonstrating its effectiveness in integrating contextual knowledge to improve emotion classification. In WisdoM, the method of converting images into text captions results in the loss of a significant amount of image information and fails to focus on the more detailed emotional expressions within the images. EMO-LLaMA \cite{xing2024emo} addresses this issue by training a dedicated encoder for facial features to extract both global and local information from faces. It employs handcrafted prompts to characterize user profiles and combines the previously extracted facial information to fine-tune the LLaMA model, thereby alleviating the model's inaccuracies in recognizing image details. This model achieved the highest accuracy on both static and dynamic facial expression recognition datasets. Emotion-LLaMA \cite{cheng2024emotion} additionally encodes audio; it uses a specific emotional encoder to seamlessly integrate audio, video, and text modalities, and performs LoRA fine-tuning on the LLaMA2 model. It achieved the highest score on the EMER dataset. Omni-Emotion \cite{yang2025omni} has a similar model structure to Emotion-LLaMA and explores various fusion schemes of facial features with other visual features. Experiments found that the method of separately extracting the visual and facial features of the entire video and then concatenating them has the least impact on MLLMs.


\subsubsection{Aspect-Based Sentiment Analysis} Aspect-Based Sentiment Analysis (ABSA) encompasses multiple subtasks, such as Aspect Term Extraction (ATE), Aspect-oriented Sentiment Classification (ASC), Joint Aspect-Sentiment Analysis (JASA), and Aspect-Category-Opinion-Sentiment (ACOS), and others. Traditional PLMs often design a separate module for each ABSA subtask, which reduces the correlation between ABSA subtasks. LLMs can leverage their strong generalization abilities to construct a unified framework for all ABSA subtasks.

Recent research explore this potential. InstructABSA introduces Instruction tuning based on the T$k$-Instruct model \cite{supernaturalinstructions} for ABSA.
This study finds that LLMs are sensitive to different instructions, with experimental results fluctuating by up to 10\% \cite{scaria2023instructabsa}.
 ITSCL \cite{zhang-etal-2024-instruction} is based on the T5 model and performs unified encoding of emotions and opinions. It introduces contrastive learning loss at each layer of the model to differentiate the feature vectors of different emotions and entity labels, thereby enhancing the model's performance. In addition to strengthening the correlations between ABSA tasks, large language models can leverage their extensive world knowledge to enhance information exchange across different domains. Some reseraches \cite{ding-etal-2024-boosting} propose an ABSA cross-domain learning fine-tuning framework. In the first stage, domain knowledge is used to train the domain variation adapter, while replay data is used to train the domain-invariant adapter, employing orthogonal constraints to force the model to learn the differences between domain-invariant and domain-variant knowledge. In the second stage, all replay data will be used to train the domain-invariant adapter while freezing the corresponding domain variation adapter. This two-stage fine-tuning effectively utilizes knowledge from different domains and avoids the issue of forgetting previous domain knowledge. The model achieved the best performance across 19 datasets.

The ABSA task shares similar objectives with sentiment analysis and sentiment polarity analysis. Unisa aims to build a unified multimodal emotion analysis model that integrates various sentiment analysis tasks \cite{li2023unisa}. This model has already established a new dataset and benchmark, SAEval, to facilitate research in this field. SAEval encompasses various ABSA sub-tasks, providing a comprehensive evaluation platform for future research.

\subsubsection{Emotion Recognition in Conversation}
\label{instruction_tuning_ERC}
Emotion Recognition in Conversation (ERC) tasks involve recognizing the sentiment label of each utterance in the conversation, which is highly dependent on the conversation history and the speaker.
InstructERC addresses these challenges by developing a template that includes a system prompt, dialogue history, and the label map. It also introduces Instruction Tuning using LoRA for ERC tasks \cite{lei2024instructerc}. Additionally, InstructERC creates two auxiliary tasks, speaker identification and emotion impact prediction, to enhance the model's understanding of speaker information. This model performs very well on the MELD \cite{poria-etal-2019-meld} and EmoryNLP \cite{zahiri2018emotion} datasets.

Researchers extend ERC tasks to incorporate multimodal inputs, considering the frequently multimodal aspects of human communication, leading to the development of Multimodal Emotion Recognition in Conversation (MERC). DialogueLLM \cite{zhang2023dialoguellm} generates text descriptions of visual information through GPT-4, combines these descriptions with the conversation history as an instruction dataset, and uses LoRA to train LLaMA2. Through experiments, it has been found that adding information from other modalities can improve the model's ability to understand dialogue information.

\subsubsection{Emotion Cause Extraction} Emotion Cause Extraction (ECE) has garnered widespread attention in recent years. This task aims to extract potential causes behind certain emotions in text, but its application scenarios are significantly limited due to the need for annotating emotional information. To address this issue, the Emotion-Cause Pair Extraction (ECPE) task propose. The ECPE task simultaneously extracts potential Emotion-Cause pairs and their corresponding causes within a document, thereby reducing the dependence on emotion labels \cite{xia-ding-2019-emotion}.
Most current research on ECPE tasks involves fine-tuning traditional PLMs \cite{ding2020end,wei2020effective} or leveraging the understanding capabilities of LLMs using well-designed prompts \cite{wu2024enhancing}. These methods primarily focus on the Textual Emotion-Cause Pair Extraction in Conversations task (TECPE). However, emotion causes may also be embedded in other modalities, leading to the development of Multimodal Emotion-Cause Pair Extraction in Conversations task (MECPE) \cite{wang2024semeval}. LLMs shows potential in better integrating information from multiple modalities for such tasks. 

Recent work on multimodal emotion‐cause understanding has progressed from clause extraction to explicit cause generation. NUS-Emo \cite{luo2024nus} attacks the original MECPE task with a two-stage self-training pipeline: LLMs are first emotion-tuned on ERC data, ImageBind aligns visual and textual affective representations, and the resulting multimodal predictions are recycled as pseudo-labels for further fine-tuning, yielding steady gains in cross-modal cause inference. Moving beyond extraction, Wang et al.\ introduce the Multimodal Emotion Cause Generation in Conversations (MECGC) task and the ObG framework \cite{wang2024observe}; an emotion-cause-aware captioner (ECCap) is trained to verbalise cause-focused video descriptions, which are then fed to an encoder–decoder generator, demonstrating that caption-based bridging markedly improves reason generation. FAME-Net \cite{wang2024generative} refines this idea by fusing holistic scene features from a visual encoder with fine-grained facial cues from VGG16, allowing dual-granularity guidance that localises both contextual and expressive triggers. Finally, Emotion-LLaMA \cite{cheng2024emotion} instruction-tunes LLaMA with a dedicated “[Reason]” token so that emotion prediction and textual justification are produced jointly, enhancing model confidence and interpretability. Collectively, these studies highlight the effectiveness of ERC pre-training, caption-based intermediaries, multi-stream visual fusion, and explanation-oriented instruction tuning for multimodal emotion-cause modelling.


\subsubsection{Subjective Text Analysis} 

LLMs are employed to explore the Stance Classification task. In this task, the model determines whether a user agrees, disagrees, or is neutral on a given point of view \cite{cruickshank2024promptingfinetuningopensourcedlarge}.
The study investigates ten open-source models and seven prompting schemes, revealing exciting conclusions. 
These include the fact that LLMs are competitive with in-domain supervised models and that the fine-tuning process only sometimes leads to better performance than zero-shot models. 
This is likely because the fine-tuning makes the model too specialized, thus unable to generalize to out-of-domain data points and varying stance definitions. 
In the opinion expression task, STOEI \cite{jia2024llm} concatenates text embeddings and speech embeddings, inputting them together into the LLM for instruction fine-tuning, enabling the model to learn to extract opinion words from the input and analyze the sentiments associated with those opinion words. This approach surpasses existing techniques by over 9.20\% and achieves state-of-the-art (SOTA) results.

Although LLMs possess inherent advantages in these subjective sentiment analysis tasks, the datasets for these tasks are mostly scarce, making instruction fine-tuning of the model challenging.

\subsection{Affective Generation}


In the current landscape of artificial intelligence, content generation that expresses sentiment is applied to various practical fields, including emotional dialogue, chatbots, emotional support conversation, and more \cite{nie2022review}. It enhances humanized services by analyzing and generating emotional information from users.

\subsubsection{Emotional Dialogue} Emotional Dialogue (ED) refers to the ability to understand and share another person's feelings and then respond in a way that conveys this understanding and compassion. LLMs must to possess high emotional intelligence to resonate with users in their replies.
However, due to the limitations of the dataset and its relative homogeneity, the generated responses often tend to be formulaic and mechanical, resulting in lower interaction quality. Some researchers aim to construct a more diverse empathetic dialogue dataset. Considering the high labor costs associated with manually constructing datasets, some current studies utilize LLM-generated synthetic data to build high-quality empathetic dialogue datasets.\cite{zheng2023building,wang2023enhancing,zhang2024stickerconv}.

To enlarge training resources for empathetic dialogue, several works exploit LLM-driven self-augmentation. SoulChat \cite{chen2023soulchat} iteratively prompts GPT-3.5-turbo to grow multi-turn exchanges, yielding richer emotional trajectories. EKTC \cite{cao-etal-2025-tool} constructs the ED-TooL dataset by asking LLaMA3 to label the exact moments at which external tools should be invoked; the resulting high-quality tool-call exemplars are then used for instruction-fine-tuning, enabling models to decide autonomously—and more cleanly—when tool usage will enhance their responses.
In the task of empathetic responses, the information dimension carried by a single text modality is relatively limited. Stickers and voice can convey non-verbal cues in interpersonal communication, and multimodal interaction mechanisms hold significant value in enhancing empathetic expression. StickerConv generates a multi-scenario, multimodal empathetic dialogue dataset constructed from user information through a multi-agent system, enabling more natural and fluid interactions with users by generating stickers within the dialogue \cite{zhang2024stickerconv}.

Speech–centric approaches further exploit prosody to boost empathy. BLSP-Emo \cite{wang2024blsp} follows a two-step alignment strategy: (i) convert the speaker’s audio into an explicit textual emotion hint and let an LLM draft a reply; (ii) apply QLoRA to inject the raw audio encodings themselves, forcing the model to ground its responses directly in vocal affect. PerceptiveAgent \cite{yan2024talk} transcribes prosodic features with an audio-encoder–text-decoder pair, feeds the resulting description together with dialogue history to an LLM, and finally vocalises the generated reply via a synthesiser—thus closing the speech-to-speech loop. EMOVA \cite{chen2024emova} generalises this pipeline to multimodal inputs and, crucially, lets the model emit style tokens that steer the synthesiser, yielding richer and more engaging emotional speech.

\subsubsection{Chatbot} 
Current datasets still limit most of the ERG and ESC tasks described above and need to capture the diversity of users and scenarios in real life fully. 
Chatbots capable of understanding user input and emotions are designed to address the practical application of emotional dialogue systems. These chatbots can offer encouragement during adversity and share in users' happiness.
PICA \cite{zhang2023PICA} and EMOLLM \cite{EmoLLM} are two notable examples of such chatbots. While LLMs can function well as "assistants," their emotional intelligence is often lacking, and long replies frequently fail to convey empathy. To address this issue, PICA is proposed. It uses P-tuning to fine-tune the ChatGLM2-6B model on two synthetic Chinese empathetic dialogue datasets, effectively enhancing the emotional capabilities of LLMs.
EmoLLM is a series of large mental health models that can support the mental health counseling link of understanding users, supporting users, and helping users. It uses the QLoRA\cite{dettmers2024qlora} method to fine-tune multiple LLMs, such as Deepseek-R1\_14b\_int4, LLaMA3-8B-Instruct, and others, across different languages. By supporting various roles, EMOLLM enables the chatbot to generate more contextually appropriate user responses.

\subsection{Multi-task of Affective Computing}
Many experiments show that LLMs possess strong comprehension and generalization capabilities. 
Furthermore, their sequence generation ability enables them to construct a unified paradigm for AU and AG tasks.
Consequently, researchers began exploring the development of a general AC model.

Due to the high similarity of mission objectives and approach in affective understanding tasks, researchers try to build a unified model that can accomplish tasks simultaneously. Emollms \cite{liu2024emollms} develops a multi-task instruction dataset, AAID, covering SA and ERC. Experimental results show improved performance on each task after multi-task training. GSA-7B \cite{hou-etal-2024-progressive} undergoes progressive LoRA training on three tasks, SA, ERC, and SD that are both related and increasingly challenging. It has been found that this hierarchical training approach can effectively improve the model's performance across multiple tasks.
In addition to classification tasks, M2SE \cite{li2024m2se} also introduces tasks such as sentiment cause extraction, achieving a unified fine-tuning framework for multimodal and multitask scenarios. By sharing model parameters and facilitating knowledge transfer between tasks, it enhances the model's ability to generalize in complex emotional contexts. This model is applicable to various emotion computation tasks.

However, current fine-tuning methods present limitations when dataset sources differ across tasks. Fine-tuning a model under these conditions can create conflicts and impact the original world knowledge embedded in the model \cite{li2024mixlora}. These challenges highlight the need for more research on ensuring high-quality performance across all sentiment analysis tasks.
To address these issues, researchers propose a novel Modular Emotional Intelligence enhancement method (MoEI) featuring two collaborative techniques—Modular Parameter Expansion (MPE) and Intra-Inter Modulation (I$^2$M)—is proposed to fit most emotion analysis tasks \cite{zhao2024both}. 
In conjunction with this method, the study develops EIBENCH, a unified affective computing dataset containing 88 datasets across 15 EI-related tasks, is introduced. MoEI adopts the Mixture-of-LoRA (MoLoRA) method for fine-tuning to improve EI while maintaining the generative ability of the model.

In multimodality, MEILLM\cite{dong2024emoada} integrates deep feature vectors extracted from visual, textual, and audio models into baichuan13B-chat through the MLP layer. After fine-tuning the instructions, he can complete a variety of emotional interaction tasks, including psychological assessment, psychological portrait, psychological counseling, etc.
MODA \cite{zhang2025modamodularduplexattention} develop a multimodal artificial intelligence that is both emotionally and intellectually capable. It addresses the issue of attention imbalance among different modalities in large models through a duplex attention alignment mechanism and a modular attention mask, allowing the model to focus more on fine-grained information from auxiliary modalities. In terms of cognitive and emotional understanding, MODA can accurately identify user intentions and emotional inclinations, demonstrating immense potential in the field of human-computer dialogue.

\subsection{Challenge}
While Instruction Tuning has enabled LLMs to make significant strides in AC tasks, several challenges remain to be addressed.
\begin{itemize}[leftmargin=*]
    \item FPFT LLMs for specific domains are time-consuming, labor-intensive, and requires significant computational resources. 
    While PEFT can reduce resource consumption in fine-tuning LLMs, they are more suitable for specific datasets and often sacrifice the generalization ability of LLMs. There is a need for scholarly research about the comparative efficacy of various PEFT methods on AC tasks.
    Consequently, developing a simple yet effective fine-tuning approach that balances efficiency, task-specific performance, and generalization capability remains a significant challenge in LLMs.
    \item While LLMs with instruction tuning demonstrate excellent performance in AC tasks, they are still slightly inferior to the previous fully fine-tuned PLMs method in some tasks. Achieving state-of-the-art performance with minimal training remains a major challenge.
    \item Despite their known universality, LLMs often struggle to simultaneously excel in both AU and AG within the domain of AC. It remains a significant challenge to develop a unified model capable of effectively addressing AU and AG.
    \item Research in niche areas of AC tasks, such as financial sentiment analysis, sarcasm detection, and metaphor analysis, still needs to be improved. Furthermore, the full potential of LLMs in these specialized AC tasks has yet to be thoroughly explored and evaluated.
\end{itemize}

%% file: _5_prompt.tex
\section{Prompt Engineering}
\label{section:prompt_engineering}


Prompt engineering~\cite{brown2020language} guides LLMs to produce desired outputs by designing appropriate prompts, effectively improving the accuracy and reliability of AC tasks. This section reviews recent advances in prompt engineering for AC, focusing on four techniques: zero-shot prompting, few-shot prompting, Chain-of-Thought (CoT) prompting, and agent.

\textbf{Zero-shot Prompting}~\cite{brown2020language} generates text for a specific topic or domain without any demonstrations. It is simple to use, highly adaptable, and especially suitable for scenarios where data are scarce or hard to obtain.


\textbf{Few-shot Prompting}~\cite{brown2020language} supplies the model with a small set of task-relevant examples or demonstrations. These examples, together with the task description, are used as prompts to help the model better understand the task, reason more effectively, and produce high-quality outputs.

 
\begin{figure*}
  \centering
    \includegraphics[width=1\textwidth]{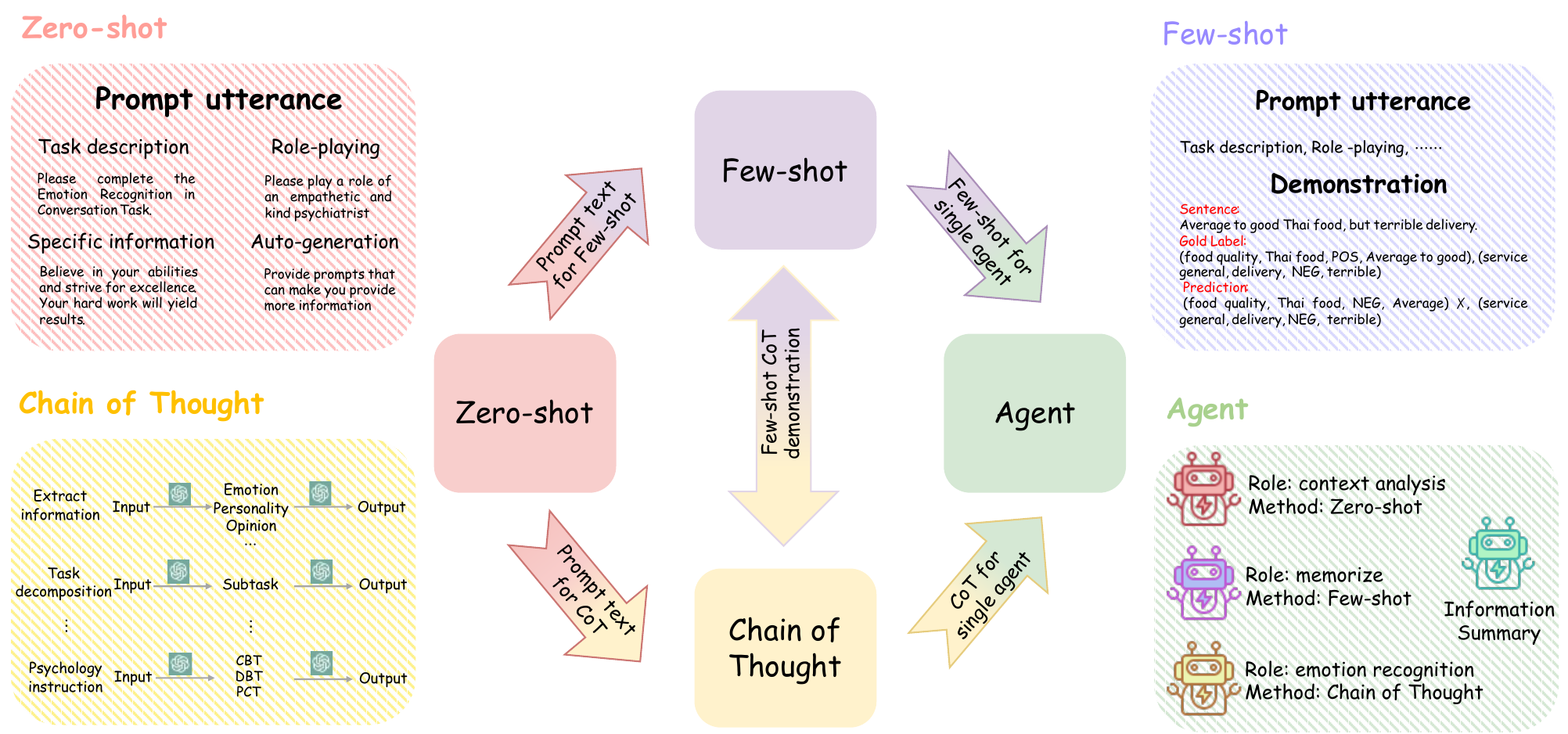}
  \caption{The overview of prompt engineering. }
  \label{fig: prompt overview}
\end{figure*}

\textbf{Chain-of-Thought}~\cite{wei2022chain} adopts a multi-step reasoning paradigm that guides the LLMs to reason logically and explicitly reveal the intermediate steps of its reasoning chain. Chain-of-Thought enables LLMs to better leverage context, produce more accurate and interpretable answers, and perform better on complex reasoning tasks.


\textbf{Agent}~\cite{park2023generative} assigns clear roles, objectives, and collaboration protocols to multiple LLMs, driving them to collaborate via message passing and feedback loops to complete tasks. Agent effectively handles task decomposition and planning, tool invocation, and memory management, and excels in open-domain, multi-stage, highly interactive scenarios.


While these prompting paradigms share commonalities, they differ in emphasis and application contexts. Zero-shot prompting directs the model with task instructions alone, without examples. Few-shot prompting appends a small number of exemplars to the instruction to align format and decision boundaries. Chain-of-thought prompting explicitly exposes intermediate reasoning steps or uses trigger phrases to elicit step-by-step thinking. Agent methods orchestrate a workflow of one or more agents and tools to support task decomposition, planning, memory, and self-verification, and to aggregate multiple outputs. Figure~\ref{fig: prompt overview} summarizes the distinctions and composability of the four approaches.


\subsection{Affective Understanding}

\subsubsection{Zero-shot for Affective Understanding}
\label{section:zero-shot for au}

In affective understanding (AU), zero-shot prompting has been applied. Researchers preset a task description as the prompt to the target text to guide LLMs to produce sentiment labels~\cite{amin2023wide,ouyang2024stability,al2023sentiment,10.1145/3603287.3651183}. Building on this, incorporating dialogue context~\cite{hong2025aer}, multi-label prompts~\cite{zhou2023prompt}, constraint-based prompts~\cite{bai2024compound}, and task-specific prefixes~\cite{cabello2024simple} further improves LLM performance on AU tasks.
Researchers explore a variety of zero-shot prompting strategies to improve the performance of LLM-based models on the AU tasks. LLMs are cast as emotion experts~\cite{wang2023enhance}, assigned specific personas~\cite{niu2024text}, or given gendered roles~\cite{plaza2024angry} to perform sentiment analysis. Psychologically inspired prompts are also introduced, such as emotional stimulation questions~\cite{li2023large} and integrating narrative style elements extracted by HEART~\cite{shen2024heart}. In Speech Emotion Recognition (SER), acoustic features are verbalized into textual descriptors~\cite{santoso2024large}, prompts are constructed in natural language~\cite{wu2024beyond}, and emotional cues in speech are distilled into prompt prefixes~\cite{liang2024aligncap}, thereby further improving recognition accuracy. Although fixed-template zero-shot prompting can be effective, its fixed content limits flexibility and diversity. Accordingly, multimodal sentiment analysis explores having LLMs generate task-specific prompts on demand and then using these prompts to guide the analysis of emotions in image–text pairs~\cite{wang2024wisdom}.

Additionally, subjective tasks such as Metaphor Recognition~\cite{mohler2016introducing} and Dark Humor Detection~\cite{hoffmann2022training} require greater attention to emotional responses. Reasoning in Conversation (RiC)~\cite{wang2024reasoning} is a zero-shot paradigm that addresses tasks via dialogue simulation, extracting key information to support the final answer.


\subsubsection{Few-shot for Affective Understanding}
\label{section:fewshot for au}

The pivotal role of demonstrations in prompt drisves exploration of the few-shot prompting. Early methods required manually curating and inserting demonstrations~\cite{zhou2024comprehensive}. With a range of selection strategies emerging~\cite{zhang2024instruction}, a central challenge is how to more effectively elicit and unlock LLMs’ few-shot capabilities for sentiment analysis.
For the challenging Aspect-Sentiment Quad Prediction (ASQP) task~\cite{zhang2021aspect}, The Self-Consistent Reasoning-based Aspect-Sentiment Quad Prediction model (SCRAP)~\cite{kim2024self} integrates pre-generated reasoning demonstrations into the task’s reasoning process to strengthen LLM reasoning. Few-shot prompting has also been introduced into multimodal sentiment analysis (MSA) for AU. WisdoM~\cite{yang2024empirical} targets four multimodal entity-based sentiment analysis (MEBSA) subtasks by converting visual inputs into textual surrogates treated as auxiliary statements, and constructing demonstrations by selecting positive and negative instances based on textual similarity. During context generation, LVLMs~\cite{liu2024visual,ye2024mplug} use prompt templates, together with the given image–sentence demonstrations, to produce contexts that explicitly incorporate world knowledge.

Additionally, in Depression Detection task, few-shot prompting guides data generation to enhancing the Structural Element Graph (SEGA)~\cite{chen2024depression}. Researchers follow principles to steer the synthesis process and provide the LLM with generation exemplars, yielding high-quality synthetic data.


\subsubsection{Chain-of-Thought for Affective Understanding}
\label{section:cot for au}

Zero-shot or few-shot prompting carries limited information in a single-turn interaction, making it insufficient for complex AU tasks. To address this, researchers introduce CoT~\cite{lee2024analyzing}, which explicitly adds intermediate reasoning steps to better handle complex tasks.
Given the complexity of AU, THOR~\cite{fei2023reasoning} employs a three-hop CoT framework for Implicit Sentiment Analysis (ISA). It sequentially extracts fine-grained aspects, analyzes latent opinions, and infers implicit sentiment. In addition, injecting causal relations as causal cues into the model~\cite{lyu2024llms} can enhance LLMs’ sentiment analysis performance.
Unlike traditional methods that extract regularities directly from text, Chain of Empathy (CoE)~\cite{lee2023chain} combines multiple psychotherapeutic approaches~\cite{beck1979cognitive,linehan1987dialectical,cooper2011person,wubbolding2017using}, designs four types of chains of thought, and interprets users’ psychological states within a professional framework, thereby enabling a more precise understanding and handling of emotions. Meanwhile, other work~\cite{manzoor2024can} adopts a psychological information–mining perspective and explores LLMs’ potential for empathetic understanding. ECR-Chain~\cite{huang2024ecr} draws on appraisal theory’s “stimulus–appraisal–emotion” process and employs multi-step reasoning to infer the deeper causes of emotions.

Methods such as CoE and ECR-Chain generate task-specific reasoning and achieve good results, but they still face challenges on complex extraction tasks such as Joint Aspect Sentiment Analysis (JASA) and Emotion-Cause Pair Extraction (ECPE). Accordingly, RP-CoT~\cite{wang2023enhance} adopts multi-step reasoning to decompose the task: it first identifies aspect-level sentiment polarities and then determines the overall polarity. Likewise, DECC~\cite{wu2024enhancing} first detects emotions and locates the corresponding emotion clauses, and then, for each emotion, selects the most probable cause clause and outputs emotion–cause pairs. Beyond generating intermediate reasoning for emotion understanding and task decomposition, CoT also supports prompt construction. The Evolutionary Multi-Objective (EMO) process~\cite{baumann2024evolutionary} starts from human-written seed prompts, feeds them to an LLM and evaluates fitness via the generated stories. It then performs crossover recombination and applies mutation operators, iteratively producing new prompts.

The CoT can be combined with few-shot prompting. For Market Sentiment Analysis~\cite{chen2018ntusd}, the complexity of financial and social-media terminology and data scarcity constrain traditional supervised methods. Researchers construct contextual exemplars and use few-shot prompts to elicit CoT summaries~\cite{deng2023llms}, prompting the model to retrieve and integrate domain knowledge before concluding and thereby injecting relevant financial knowledge into the Market Sentiment Analysis.


\subsubsection{Agent for Affective Understanding}
\label{section:agent for au}

Prior studies mostly adopt a single LLM decision paradigm~\cite{fei2023reasoning,yang2024empirical}. However, a single model’s output does not reliably approach the optimum. To address this, Agent techniques orchestrate collaboration and  evaluation among multiple LLMs, enabling more robust ensemble decisions.

To achieve multi-LLM negotiation in sentiment analysis, the Generator-Discriminator Role-switching Decision-Making framework~\cite{sun2023sentiment} is proposed. The generator injects reasoning and provides decisions with rationales, the discriminator explains and assesses credibility, and the two iterate until they reach consensus, leveraging model complementarity for more accurate emotion interpretation and correction. Building on this idea, agent design expands with an automated 5W1H question-generation module to predict emotions from social information~\cite{wei2024mimicking}. Multi-agent interaction also builds an emotion contagion model that characterizes how emotions propagate within groups and their effects~\cite{van2023empirical}. The PANAS framework~\cite{regan2024can} employs three specialized agents for memory, context understanding, and evaluation. Meanwhile, AnnaAgent~\cite{wang-etal-2025-annaagent} uses multi-session memory to capture users’ long-term emotional dynamics, offering a practical route to more realistic and deeper digital companions.


In Financial Sentiment Analysis (FSA), HAD~\cite{xing2024designing} summarizes five common error types and, accordingly, instantiates five specialized agents—sentiment, rhetoric, dependency, aspect, and reference. Each agent targets the error-prone dimension for LLMs, and their outputs are aggregated and deliberated to produce the final FSA decision.


\subsection{Affective Generation}

Prompt engineering is also used to improve the performance of AG. The following introduces the application of four prompt engineering technologies in AG.


\subsubsection{Zero-shot for Affective Generation}
\label{section:zero-shot for ag}

To align prompts with the characteristics of AG, researchers propose Perspective-Taking~\cite{lee2023investigating}, which enhances LLMs’ empathetic dialogue generation by shifting viewpoints. In HEF~\cite{yang2024enhancing}, a small empathy model first extracts emotion causes and primary emotion categories. These are then provided to the LLM under zero-shot prompting to produce more precise empathetic responses. The TCG pipeline~\cite{bhaskar2022prompted}, built on GPT-3.5-turbo, performs sentence-level opinion extraction, grouping, and clustering to generate long-form summaries of affective opinions.


In the emotion-conditioned text generation task, researchers optimizes prompts by iteratively adding, replacing, and deleting the original prompts~\cite{resendiz2023emotion}. The optimized prompts better achieve the expected emotions, resulting in improved emotional text generation. Since the output of LLMs largely depends on prompts, users needing more background knowledge might experience a diminished interaction with chatbots due to the absence of contextual understanding. To address this issue, PromptMind~\cite{su2023prompt} generates multiple contextually relevant prompts during the conversation for users to choose from, thereby prompting LLMs to produce answers that better meet user expectations and enhance human-chatbot interaction.

\subsubsection{Few-shot for Affective Generation}
\label{section:fewshot for ag}

In the Sibyl framework~\cite{Wang2023SibylSE}, researchers randomly select a single example as a demonstration to prompt the LLM to generate four types of prospective commonsense related to the dialogue, which are then used as training data for subsequent dialogue-generation models. For mixed-initiative dialogue generation tasks such as PersuasionForGood and Emotional Support Conversations, responses are generated conditioned on different exemplar strategies~\cite{chen2023controllable}.


Researchers indicate that generated responses utilizing prompt engineering demonstrate high quality and adherence to semantic controls~\cite{chen2023controllable}. However, in emotional support tasks, the inherent preference of LLMs for specific strategies can hinder effective support despite improving the robustness of predicting appropriate strategies~\cite{kang2024can}. To address this issue, researchers conduct extensive experiments involving Self-contact approaches, external strategy planning, and example expansion, demonstrating that External-contact approaches help reduce LLM preference bias and enhance the performance of emotional support.


\subsubsection{CoT for Affective Generation}
\label{section:cot for ag}

In AG, information and knowledge are crucial for producing high-quality responses. To obtain more valuable information, researchers introduce CoT to support empathetic generation. CFEG~\cite{chen2024cause} decomposes emotion and reasoning into multiple steps, strengthening the listener’s role awareness and significantly improving the empathy of LLMs. User utterances serve as latent linguistic cues that reveal hidden needs and guide the model to generate more personalized responses. Cue-CoT~\cite{hongru2023cue} breaks down reasoning into sequential steps to infer user states (e.g., personality, psychological status, emotions) and finally generates responses based on the reasoning outputs. EDIT~\cite{wu2023new} enhances response generation by detecting implicit user intentions. It first asks open-ended questions to capture potential intentions, then retrieves knowledge base answers, and finally integrates the additional knowledge into generation. Aptness~\cite{hu2024aptness} adopts a similar strategy by constructing an empathetic response database via multi-step CoT and using retrieval during generation to enhance empathetic capability.


Since AG requires LLMs to address emotion-related issues more proactively, planning information plays an important role in the process of obtaining and using information with LLMs. ProCoT~\cite{deng2023prompting} guides LLMs to produce intermediate steps of reasoning and planning and analyzes through dynamic reasoning how to achieve dialogue goals, which enhances proactive dialogue ability. ECoT~\cite{li2024enhancing} proposes five consecutive steps to simulate human emotional intelligence and generate responses with richer emotional expression. ESCoT~\cite{zhang2024escot} promotes response generation by producing emotional stimuli and designing regulation strategies. Further research transforms step-by-step cognitive commentary prompts into a hierarchical concept~\cite{li2024sentiment} and proposes a three-level emotional integration framework.


Additionally, in metaphor generation, GROUNDS naturally serve as the reasoning chain that connects TENOR and VEHICLE. Researchers guide the model to generate logically consistent GROUNDS~\cite{shao2024cmdag}, which in turn produce coherent and rich VEHICLE expressions, thereby improving the quality of metaphor generation.


\subsubsection{Agent for Affective Generation}
\label{section:agent for ag}

As a new technology based on LLMs, agent is gradually applied to AG. The Agent4SC~\cite{zhang2024stickerconv} generates user information, enhances memory, and formulates response strategies through collaboration among LLMs, realistically simulating human sticker usage to improve multimodal empathetic dialogue. The COOPER dialogue framework~\cite{cheng2024cooper} coordinates multiple specialized agents focusing on specific dialogue goals to generate emotionally supportive and persuasive responses. Researchers propose a multi-turn dialogue framework based on psychologist agents~\cite{wu2024llm}, in which debaters from different psychological schools generate candidate replies, and an unbiased decision-maker selects the final response, thus addressing the challenge of integrating multiple psychological schools in LLMs. The PerceptiveAgent~\cite{yan2024talk} simulates different speaking styles according to prompts to build multimodal dialogue systems. The conversational agent with acoustic emotion perception (CA)~\cite{hu2022acoustically} performs speech emotion recognition and incorporates empathetic feedback and interjections to give responses an emotional style. In addition, introducing self-emotion prompts is also shown to help Agents demonstrate more human-like dialogue strategies~\cite{zhang2024self}.

Additionally, in the multimodal domain, researchers study multimodal Conversational Health Agents (CHAs)~\cite{abbasian2024empathy} and design five types of agents as components to interact with users to generate audio responses based on user emotions. 

\subsection{Challenge}

Despite the significant achievements of Prompt Engineering in the field of AC, the following challenges persist:
\begin{itemize}[leftmargin=*]
    \item As an efficient method with low resource requirements, it greatly reduces the cost of models. However, its performance indicators across various tasks still need to catch up to those of fine-tuned models, unable to meet practical application needs.
    \item Currently, the most widely used and advanced CoT method has significantly improved the effectiveness of AC tasks. However, its underlying logic aims to fit human reasoning processes, which conflicts with the unique mechanisms of human emotion generation. This often results in noise during the intermediate reasoning process, affecting the final inference outcomes.
    \item Moreover, since emotions are influenced by various factors such as psychology, knowledge, and context, the ways these factors impact and are obtained vary greatly. The current division of agents is not detailed enough, each agent's design is overly simplistic, and the information aggregation process is quite arbitrary, failing to fully exploit the potential of LLMs.
\end{itemize}

%% file: _6_reinforcement.tex
\section{Reinforcement Learning}
\label{section:rl}

\begin{figure*}
\centering
\includegraphics[width=0.85\linewidth]{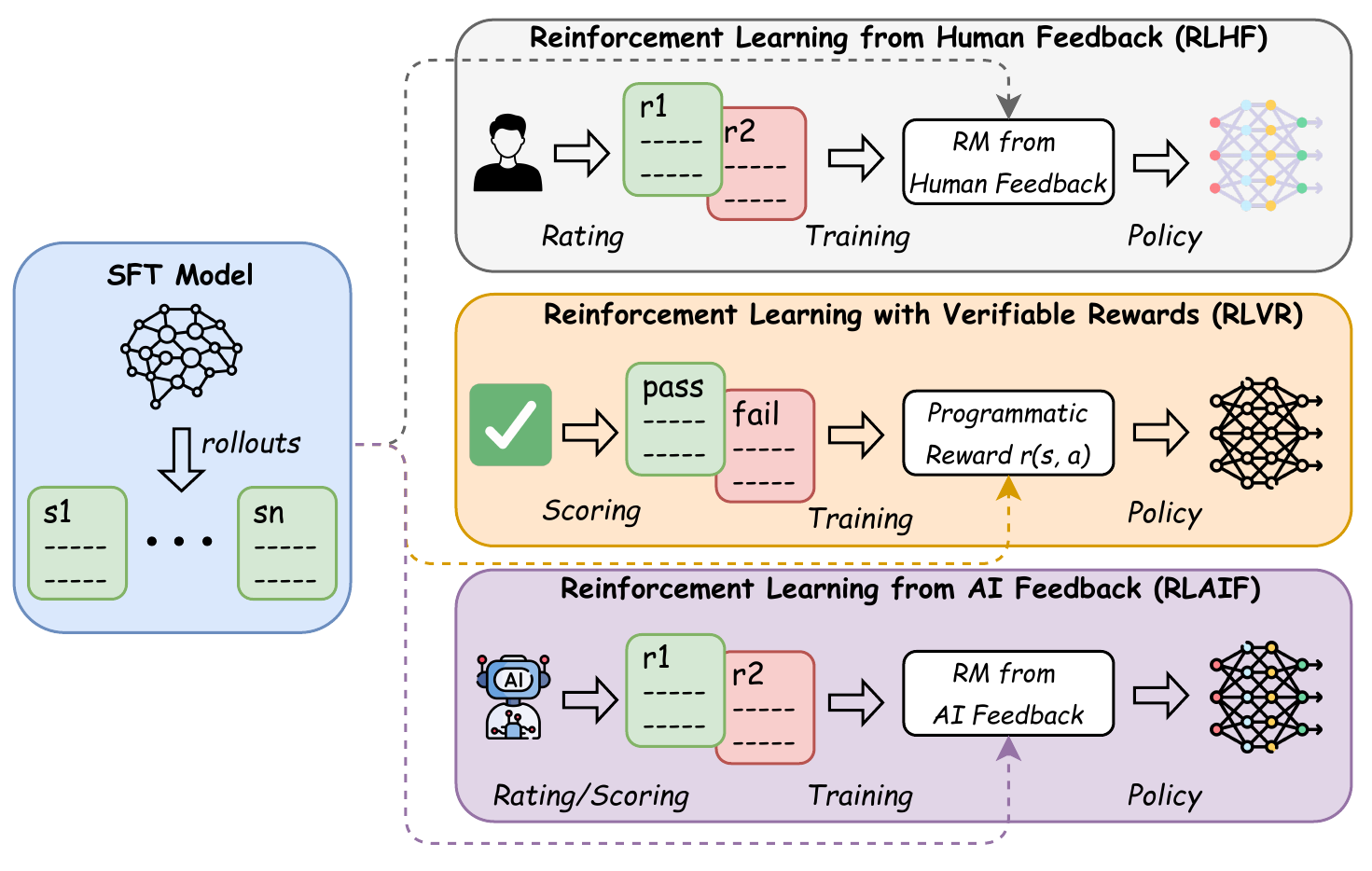}
\caption{The overview of reinforcement learning.}
\label{fig:rl-layout}
\end{figure*}

Reinforcement learning (RL) is increasingly used to align large language models (LLMs) with affect-related objectives beyond supervised fine-tuning. We organize RL for AC into three families that differ by the source of rewards yet share the goal of optimizing behavior at token-, utterance-, and dialogue-levels across AU/AG (see Figure~\ref{fig:rl-layout}): (i) \emph{Reinforcement Learning from Human Feedback (RLHF)}~\cite{ouyang2022training, rafailov2023direct}, where a reward model trained on human comparisons guides policy optimization—often with \emph{Proximal Policy Optimization (PPO)}—and preference-only variants such as \emph{Direct Preference Optimization (DPO)} reduce reliance on online RL~\cite{ouyang2022training,schulman2017proximal,rafailov2023direct}; (ii) \emph{Reinforcement Learning with Verifiable/Programmatic Rewards (RLVR)}~\cite{lambert2024tulu}, which replaces preference labels with rule- or checker-based signals and has inspired algorithms like \emph{Group Relative Policy Optimization (GRPO)}~\cite{shao2024deepseekmathpushinglimitsmathematical} for stable group-wise ranking~\cite{lambert2024tulu,shao2024deepseekmathpushinglimitsmathematical}; and (iii) \emph{Reinforcement Learning from AI Feedback (RLAIF)}, which leverages capable AI evaluators to provide multi-aspect feedback that can be plugged into RL or preference-optimization pipelines. The following subsections detail representative methods and applications of these three families in affective computing.

\subsection{Reinforcement Learning from Human Feedback}
\label{section:rlhf for rl}

A key application of RLHF is enhancing empathy in dialogue systems. Zhu et al.~\cite{zhu2023grafting} propose a three-stage framework for Empathetic Emotion Elicitation, where reinforcement learning leverages user reactions as rewards to improve emotional engagement across conversations.
However, collecting human feedback is costly and difficult to scale. To mitigate this, Xu et al.~\cite{xu2025rlthf} introduce Reinforcement Learning from Targeted Human Feedback (RLTHF), which first uses a general-purpose LLM to label data, then directs limited human effort toward ``hard'' or mislabeled cases. This approach achieves full human-level alignment at only 6--7\% of the annotation cost.

The quality of preference data remains critical. Liang et al.~\cite{liang2024aligncap} present AlignCap, a preference optimization method that aligns speech emotion captions with human preferences, reducing hallucinations and producing more faithful emotional descriptions. Complementarily, Chen~\cite{chen2025empathyagent} release \emph{EmpathyAgent}, the first benchmark for training and evaluating empathetic embodied agents, offering 10,000 multimodal samples with preference-labeled empathetic actions for RLHF training.

\subsection{Reinforcement Learning with Verifiable Rewards}
\label{section:rlvr for rl}

To bypass the human annotation bottleneck, RLVR leverages programmatic or model-based reward functions, enabling scalable and automated optimization.
A representative work is RL-EMO by Zhang et al.~\cite{zhang2024rl}, the first to apply RL to multimodal emotion recognition in conversation (ERC). It frames ERC as a sequential decision-making problem, where correct or incorrect emotion predictions yield positive or negative rewards, respectively. Using a Deep Q-Network to model emotion flow, RL-EMO outperforms prior methods and demonstrates the effectiveness of policy optimization for capturing temporal affective dynamics.
Zhang et al.~\cite{zhang2025sage} further propose SAGE (State Augmented GEneration), a framework that explicitly models user and assistant emotional states as latent variables for dialogue generation. Reinforcement learning is applied at the state level, enabling strategic emotional control and producing more empathetic and context-aware responses.
Extending into the multimodal domain, Zhao et al.~\cite{zhao2025r1} present R1-Omni, which applies RLVR to omni-multimodal (video, audio, text) emotion recognition. With a simple verifiable reward—whether predictions match ground-truth labels—optimized via Group Relative Policy Optimization (GRPO), the framework improves accuracy, robustness, and interpretability in emotion reasoning.

Pushing the boundaries of affective understanding beyond static emotion recognition, the SAGE (Sentient Agent as a Judge) framework~\cite{zhang2025sentient} reconceptualizes comprehension as a dynamic, interactive process. Its core innovation is a "Sentient Agent" that simulates human-like emotional changes and inner thoughts during multi-turn conversations. By embedding evaluation within the agent’s evolving persona, goals, and emotional state, comprehension is reframed from classification to a stateful simulation of a user "feeling understood." The resulting emotion trajectories and interpretable inner thoughts provide a proxy for modeling and measuring the essence of being heard.

Most RL applications in affective computing focus on generation, steering LLMs toward empathetic, polite, or strategically effective outputs. RLHF naturally fits this domain by embedding human preferences into the training loop. For example, RLVER by Wang et al.~\cite{wang2025rlver} introduces the first end-to-end RL framework leveraging verifiable emotion rewards from simulated users. Built upon the SAGE simulator, it produces deterministic emotion scores as direct PPO rewards, elevating a 7B open-source model to near-proprietary performance on Sentient-Benchmark.
A key innovation in RLVR is finer-grained reward design. Li et al.~\cite{li2024reinforcement} propose TOLE, which formulates token-level rewards via classifier probability shifts, enabling efficient controllable text generation. Cai~\cite{cai2024empcrl}’s EmpCRL combines empathy and diversity signals, while Li et al.~\cite{li2024helpful} integrate the "Cognitive Relevance Principle" into ORL, balancing helpfulness against user effort. Together these works show how automated, dense rewards lead to nuanced affective generation.
Ma et al.~\cite{ma2025empathy} propose EmpRL, aligning response empathy levels through a programmatic reward from a dedicated "empathy identifier," based on reaction, interpretation, and exploration. Chang et al.~\cite{chang2024applying} further demonstrate multi-objective reward design in emotional support systems, combining goal-level, flow, and valence rewards for stage-aware dialogue policies.
RLVR also supports domain-specific agents. Mishra et al.~\cite{mishra2024able} introduce ABLE, a disability support agent guided by a six-component reward model covering persona consistency, politeness, empathy, naturalness, and coherence. Priya et al.~\cite{priya2025genteel} develop GENTEEL-NEGOTIATOR, which uses MoE-based RL with detailed reward functions for polite yet strategic negotiation. For online safety, Wang et al.~\cite{wang2024f2rl} present F$^2$RL, leveraging triplet-based factuality and multi-aspect faithfulness rewards to generate effective counterspeech. 

Beyond dialogue, RLVR principles extend into strategic decision-making. Unnikrishnan et al.~\cite{unnikrishnan2024financial} apply sentiment-aware RL to portfolio management, where sentiment scores from financial news served as programmatic rewards. The resulting trading agent achieved superior profitability, showing how effective signals can optimize complex, non-dialogue tasks.

\subsection{Reinforcement Learning from AI Feedback}
\label{section:rlaif for rl}
Within the RLVR paradigm, Reinforcement Learning from AI Feedback (RLAIF) replaces human annotators with capable AI models to provide reward signals. Yoshida et al.~\cite{yoshida2025training} exemplify this approach by training an LLM-based reward model on 12 conversational metrics (e.g., empathy, consistency, enjoyability). This AI-generated, multi-faceted feedback is then used with DPO to optimize dialogue agents, showing that AI feedback can scalably capture holistic qualities beyond turn-level scoring.

RLVR also supports new agent architectures. Yuan~\cite{yuan2024reflectdiffu} propose ReflectDiffu, integrating RL into diffusion models where a policy network selects empathetic intents conditioned on emotional contagion and mimicry. High-level planning approaches include Deng~\cite{deng2023plug}'s Plug-and-Play Policy Planner and Rakib~\cite{rakib2025dialogxpert}'s DialogXpert, which employ LLM critics in self-play to train efficient policy planners for proactive tasks such as emotional support and negotiation.
The paradigm further extends to autonomous self-improvement. Ye et al.~\cite{ye2025generic} propose a self-evolution framework where models refine their own outputs to generate synthetic preference pairs for DPO. Zhao et al.~\cite{zhao2025chain} advance this idea with Chain-of-Strategy Optimization (CSO), which uses Monte Carlo Tree Search to explore conversational strategies. Candidate responses are scored by an auxiliary reward model, yielding high-quality preference datasets for DPO fine-tuning, significantly improving strategic empathy.
Zhang et al.~\cite{zhang2025decoupledesc} introduce Decoupled ESC, featuring Inferential Preference Mining (IPM) that identifies psychological errors (e.g., lack of empathy, premature shifts) in model outputs. Erroneous responses are paired with ground-truth ones to build synthetic preference data. The resulting decoupled system, with separate strategy planning and response generation, yields more robust and effective emotional support agents. Collectively, these works illustrate the trajectory of RLAIF toward fully autonomous systems capable of continuously enhancing emotional intelligence.

\subsection{Challenge}
\label{section:ch for rl}
Despite the growing success of reinforcement learning in aligning LLMs with affective objectives, several key challenges remain:

\begin{itemize}[leftmargin=*]
    \item While RLHF has significantly improved model alignment with human preferences, it remains highly resource-intensive, requiring large-scale, high-quality human preference data. This poses a major scalability bottleneck, especially for fine-grained affective tasks that require nuanced and subjective judgments.
    \item Many current reward functions used in RLHF and RLVR are designed based on task-level or surface-level attributes (e.g., helpfulness, politeness), but they often lack grounding in psychological or cognitive theories of emotion. This can lead to misalignment between the reward signal and the underlying affective dynamics, limiting the depth of emotional understanding or expression the model can achieve.
    \item Despite advances in RL for affective understanding (e.g., RL-EMO, SAGE), temporal affect modeling remains underexplored. Emotions evolve over time, yet most RL-based affective systems still optimize at the turn-level or token-level, failing to capture long-term emotional trajectories in conversation.
    \item The lack of standardized benchmarks and evaluation metrics for affective RL makes it difficult to systematically compare approaches or track progress. Current evaluations often rely on subjective human scoring or limited automatic proxies, which may not fully reflect emotional richness, coherence, or alignment with human expectations.
\end{itemize}

%% file: _7_benchmark.tex
\section{Benchmark \& Evaluation}
\label{section:benchmark}
This section compiles various benchmarks, evaluation methods, and metrics for Affective Computing, encompassing both Affective Understanding and Affective Generation. Table \ref{Table:benchmarks} comprehensively summarizes these benchmarks, detailing the evaluated tasks, datasets, metrics, and models.

\begin{table*}[htbp]
\renewcommand\arraystretch{1.2}
    \centering
    \caption{Various benchmarks for Affective Computing. The AE means Automatic Evaluation, the HE means Human Evaluation.}

    \resizebox{\textwidth}{!}{%
    \begin{tabular}{lcccc}
    \toprule[1pt]
        \textbf{Name} & \textbf{Tasks} & \textbf{Datasets} & \textbf{Metrics} & \textbf{Models} \\ 
    \midrule[1pt]

    SOUL~\cite{dengSOULSentimentOpinion2023}  & Sentiment Analysis & Yelp~\cite{NIPS2015_250cf8b5}, IMDb~\cite{maas-etal-2011-learning}  & AE (F1, BLEU, ...), HE & Flan-T5, GPT-3.5-turbo\\
    GPT-4V with Emotion~\cite{Lian2023GPT4VWE} & Sentiment Analysis & MVSA-Single, DFEW, etc. 21 datasets & AE (Accuracy, WAR, WAF) & GPT-4V \\
    MERBench~\cite{Lian2024MERBenchAU} & Multimodal Sentiment Analysis & MER2023, etc. 7 datasets & AE (WAF, MSE) & PLMs, LLMs \\
    
    DFEW ~\cite{jiang2020dfew} & Videol Sentiment Analysis & 1500 movies &  AE(UAR,WAR)   & EC-STFL \\
    
    MC-EIU ~\cite{Liu2024EmotionAI} & Emotion and Intent Joint Understanding in Multimodal Conversation & MC-EIU test &  AE (WAF) & EI$^2$, other PLMs \\

    RLVER~\cite{wang2025rlver} & Empathetic Understanding & Sentient Benchmark & AE (Sentient Benchmark) & GPT-4, GPT-4V, DeepSeek, Claude3.7 \\
    
    SAGE~\cite{zhang2025sentient} & Empathetic Understanding & 100-scenario Supportive-Dialogue benchmark  & AE(BLRI, Sentient emotion score) & GPT-4o, Gemini2.5\\

    \midrule[1pt]

    Cue-CoT~\cite{hongru2023cue} & Emotional Support Conversation  & ED, etc. 6 datasets & AE (Avg BLEU and F1), HE & ChatGLM, other LLMs \\

    ESC-Eval\cite{zhao2024esc} & Emotion Support Conversations & 2,801 Role Cards Benchmark &  AE (ESC-RANK)\& HE (7 dimensions) & GPT-4, Llama3, EmoLLM, etc. \\
    EmotionBench~\cite{Huang2023EmotionallyNO} & Empathetic Response Generation  & EmotionBench test & AE, HE &  GPT series, other LLMs \\
    
    STICKERCONV~\cite{zhang2024stickerconv} & Multimodal Empathetic Response Generation & STICKERCONV & AE(MMr, BLEU, ...), HE & PEGS, ChatGLM3, Vicuna \\

    SECEU~\cite{Wang2023EmotionalIO} & Emotional Intelligence Evaluation & SECEU test & AE (SECEU score, EQ score, ...) &  GPT series, and other LLMs \\
    EQ-Bench~\cite{paechEQBenchEmotionalIntelligence2024} & Emotional Intelligence Evaluation & EQ-Bench test &  AE (EQ-Bench scores) & GPT series, other LLMs  \\ 
    EmoBench~\cite{Sabour2024EmoBenchET} & Emotional Intelligence Evaluation & EmoBench test & AE (EmoBench scores) &  GPT series, other LLMs \\
    EIBENCH~\cite{zhao2024both}& Emotional Intelligence, General Intelligence & MMLU, etc. 88 datasets & AE (Accuracy) & Flan-T5, LLaMA-2 \\

    EmoVIT~\cite{Xie2024EmoVIT}& Image Sentiment Explaination Generation & EmoVIT test & AE (Accuracy) & EmoVIT, other MLLMs \\

    EMER~\cite{lian2023explainable}& Explainable Multimodal Emotion Recognition & MER2023 & AE(Accuracy, BLEU, ROUGE)& EMER model, other MLLMs \\   
    
    MER-UniBench~\cite{lian2025affectgpt}& Multimodal Emotion Recognition and Reasoning & MER-Caption & AE(UAR,WAR), HE & AffectGPT, other MLLMs \\
    
    \midrule[1pt]
    ChatGPT2AC~\cite{aminWideEvaluationChatGPT2023} & Affective Computing & res14, et 13 datasets &  AE(Accuracy, UAR) & GPT-4 \\
    
    MM-INSTRUCTEVAL~\cite{Yang2024MMInstructEvalZE} & Multimodal Reasoning with Multimodal Contexts & MVSA-Single etc. 16 datasets & AE (Accuracy) & GPT series, other LLMs \\
    \bottomrule[1pt]
    \end{tabular}
    
}
\label{Table:benchmarks}
\end{table*}

\subsection{Affective Understanding}
\label{section:au for benchmark}
This section introduces the benchmarks and representative datasets in the field of Affective Understanding for different sentiment analysis-related tasks.

\textbf{SOUL}~\cite{dengSOULSentimentOpinion2023} introduces two subtasks: Review Comprehension (RC) to validate subjective statements from reviews, and Justification Generation (JG) to explain sentiment predictions.

\textbf{GPT-4V with Emotion}~\cite{Lian2023GPT4VWE} evaluates GPT-4V's capabilities in Generalized Emotion Recognition (GER), providing quantitative results across 21 benchmark datasets spanning six visual and multimodal emotion recognition tasks.

\textbf{MERBench}~\cite{Lian2024MERBenchAU} is a benchmark for Multimodal Sentiment Analysis, covering seven prominent datasets. It uses Automatic Evaluation with metrics like weighted average F-score (WAF) and mean squared error (MSE).

\textbf{DFEW}~\cite{jiang2020dfew} is a large-scale dataset for dynamic facial expression recognition, containing over 16,000 video clips from movies with real-world challenges like occlusions and variable poses. Evaluation metrics include UAR and WAR.

\textbf{MC-EIU}~\cite{Liu2024EmotionAI} is a multimodal conversational dataset with nearly 5,000 video clips from bilingual TV series. It targets joint emotion and intent understanding in conversations, evaluated using weighted average F-score (WAF).

\textbf{RLVER}~\cite{wang2025rlver} is a reinforcement learning framework for enhancing empathetic reasoning in LLMs. It uses the Sentient Benchmark to evaluate a model's ability to improve a simulated user's emotional state over time through verifiable emotion rewards.

\textbf{SAGE}~\cite{zhang2025sentient} is an evaluation framework for higher-order social cognition in LLMs. It uses a simulated Sentient Agent to track emotional changes during dialogues, evaluating a model's ability to foster positive emotional shifts via its primary metric, the Sentient emotion score.

\subsection{Affective Generation}
\label{section:ag for benchmark}
This section introduces the benchmarks in the field of Affective Generation (AG), covering Emotional Support Conversation and Empathetic Response Generation.

\subsubsection{Emotional Support Conversation}
Emotional support conversation (ESC) aims to alleviate individuals’ emotional intensity and provide guidance for navigating personal challenges through engaging dialogue. \textbf{Cue-CoT}~\cite{hongru2023cue} is a benchmark consisting of six in-depth dialogue datasets in both Chinese and English, considering three aspects of user statuses: personality, emotions, and psychology. It forms a comprehensive evaluation benchmark for dialogue response generation, employing both Automatic Evaluation and Human Evaluation.

\textbf{ESC-Eval} \cite{zhao2024esc}  is a framework for evaluating Emotion Support Conversation (ESC) in LLMs. It uses a specialized role-playing agent (ESC-Role) and a benchmark of 2,801 role cards to simulate multi-turn dialogues. Evaluation is performed via manual annotation across seven dimensions, such as fluency and empathy, enabling a dynamic assessment that moves beyond static, ground-truth-reliant metrics.

\subsubsection{Empathetic Response Generation}
\textbf{MEDIC}~\cite{Zhu2023MEDICAM} is a multimodal empathy dataset for psychotherapy scenarios, considering visual, audio, and textual modalities. It proposes a conceptual framework for empathy with three mechanisms: Expression of Experience (EE), Emotional Reaction (ER), and Cognitive Reaction (CR).

\textbf{EmotionBench}~\cite{Huang2023EmotionallyNO} collects 428 distinct situations categorized into 36 factors and is designed to align LLM emotional responses with human appraisal, focusing on eight negative emotions.

\textbf{STICKERCONV}~\cite{zhang2024stickerconv} is a multimodal empathetic dialogue dataset with 2K generated personalities, 12.9K dialogue sessions, and 5.8K unique stickers. It generates textual empathetic responses and outputs stickers as a visual modality. Evaluation is conducted using Automatic Evaluation, Human Evaluation, and LLM-based Evaluation.

\subsubsection{Emotional Intelligence}
In the context of large language models (LLMs), emotional intelligence (EI) has emerged as a critical area of focus. EI encompasses the capacity to identify, comprehend, regulate, and apply emotions across diverse situations~\cite{mayer2016ability}. 

\textbf{SECEU (Situational Evaluation of Complex Emotional Understanding)}~\cite{Wang2023EmotionalIO} is a standardized EI test suitable for both humans and LLMs. It evaluates the ability to comprehend and apply emotions in diverse situations.

\textbf{EQ-Bench}~\cite{paechEQBenchEmotionalIntelligence2024} is a benchmark designed to evaluate aspects of EI in LLMs. The test dataset is constructed using GPT-4 to generate dialogues that serve as the context for the test questions.

\textbf{EmoBench}~\cite{Sabour2024EmoBenchET} is a theory-based comprehensive EI benchmark for LLM evaluation, consisting of 400 hand-crafted questions available in English and Chinese. The framework defines machine EI in terms of emotional understanding and emotional application.

\textbf{EIBENCH}~\cite{zhao2024both} is a large collection of EI-related tasks in a text-to-text format, spanning 15 tasks across 88 datasets. It evaluates three key EI aspects (Emotion Perception, Cognition, and Expression) and general intelligence.

\textbf{EmoVIT}~\cite{Xie2024EmoVIT} is the first instruction-tuned benchmark for visual emotion understanding, connecting visual cues with emotional cognition. It features a large-scale dataset of image-instruction pairs for tasks including classification, interpretation, and emotion-grounded conversation.
 
\textbf{EMER}~\cite{lian2023explainable} is a benchmark for Explainable Multimodal Emotion Recognition. It requires models to provide explanations for predicted emotions, evaluating their ability to integrate information from audio, video, and text modalities.

\textbf{MER-UniBench}~\cite{lian2025affectgpt} is the largest benchmark for emotion recognition, with 115K human-verified samples across over 2K fine-grained categories. It supports tasks like fine-grained/basic emotion recognition and sentiment analysis.

\subsection{Others}
\label{section:others for benchmark}
This section introduces other benchmarks that do not easily fit into AU or AG tasks, including affective computing and multimodal reasoning benchmarks.

\textbf{ChatGPT2AC}~\cite{aminWideEvaluationChatGPT2023} evaluates InstructGPT and GPT-4 across a wide range of affective computing tasks, including sentiment analysis, toxicity detection, and personality assessment. The benchmark uses Unweighted Average Recall (UAR) and Accuracy as its automatic evaluation metrics.

\textbf{MM-INSTRUCTEVAL}~\cite{Yang2024MMInstructEvalZE} establishes a benchmark for 31 models, including 23 MLLMs, using ten instructions across 16 diverse multimodal reasoning datasets with vision-text contexts. Evaluation metrics include Best Performance, Mean Relative Gain, Stability, and Adaptability.

%% file: _8_future.tex
\section{Discussion and Future Works}
\label{sec:future}
\begin{enumerate}[leftmargin=*]
    \item \textbf{Ethics}

   Data is essential and invaluable in artificial intelligence. Sentiment data from the real world, such as data from social media, psychological counseling, is highly personal, sensitive, regional, and cultural. Therefore, ethics must be considered when constructing a dataset, including individual privacy protection, prejudice elimination, and value alignment. Acquiring, preprocessing, and annotating data are all very time-consuming and labor-intensive processes. Thus, leveraging the world knowledge and generative capabilities of LLMs to synthesize datasets for affective computing tasks is an important research direction. Furthermore, improving the quality of original datasets and balancing the new and old datasets is a problem worth studying.


    \item \textbf{Multimodal Affective Computing} 
    
    Large Language Models (LLMs), initially designed to work exclusively with text, have now evolved into Multimodal Large Language Models (MLLMs), capable of processing both text and image modalities. In the domain of AC, data from diverse modalities—such as text, visuals, audio, electroencephalogram (EEG), and electrooculogram (EOG)—can convey individual emotions and sentiments. However, most current general-purpose MLLMs face limitations in their ability to integrate and interpret information from multiple modalities in a seamless manner. This often results in misinterpretations, particularly when critical emotional cues are absent or overlooked.
    To advance AC, it is essential to effectively collect and utilize multimodal data to train unified MLLMs. Key challenges include recognizing micro-expressions, capturing sentiment from audio and video, and enhancing brain-computer interface technologies and emotional intelligence systems. Addressing these challenges represents crucial research directions for the future.
    
    \item \textbf{Multilingual and Multicultural Affective Computing} 

    Multilingual and Multicultural sentiment analysis spans different languages and cultures, focusing on sentiment recognition and analysis. Users from different linguistic or cultural backgrounds express emotions differently; for example, the same body movements can have different meanings in various cultural contexts. Current sentiment analysis is typically conducted on a single language or culture. Advances in multicultural sentiment analysis will significantly enhance the understanding of global users' emotional reactions and behavior patterns, thereby improving the quality of cross-cultural communication and business decision-making. 
    When training the current LLMs, the quality and quantity of corpora in different languages vary greatly, resulting in uneven AC performance in different languages. 
    The core challenge in this field is effectively processing and integrating information from multiple languages and cultures to achieve accurate sentiment analysis.  Therefore, it is necessary to continue to improve the multilingual and multicultural capabilities of the LLMs from multiple perspectives.

    \item \textbf{Real-time Affective Computing}

    The first step in sentiment analysis is building a dataset and performing sentiment analysis. Real-time sentiment analysis in a dynamic environment, by integrating data in real-time and identifying and analyzing sentiment, is particularly critical for various practical applications such as online customer service, real-time monitoring, and interactive entertainment. Real-time sentiment analysis requires collecting and processing data in real-time, which presents several challenges—ensuring the real-time performance of data processing and sentiment analysis, especially maintaining time synchronization and consistency across different modalities.
    Real-time systems often encounter various environmental noises. Another major challenge is ensuring that the system can adapt to environmental changes and noise interference while maintaining the accuracy and stability of sentiment recognition.

    \item \textbf{Reasoning in Affective Computing}
    
    Current affective computing tasks primarily focus on understanding emotions and generating emotion-related content. However, they often overlook the deeper, underlying causes that trigger emotional states—specifically, the reasons behind the emergence, persistence, or fluctuation of certain emotions in response to various stimuli or contexts. Gaining insight into these triggers is essential for advancing systems beyond surface-level emotion recognition toward a more nuanced and comprehensive understanding of human affect.
    A major challenge lies in the extraction and integration of meaningful emotional cues to enable accurate sentiment reasoning. Successfully addressing this challenge could significantly enhance the interpretability and reliability of affective computing systems. However, evaluating and validating the sentiment reasoning remains a significant obstacle, requiring innovative approaches to balance technical accuracy with user comprehension.

    \item \textbf{Effective Evaluation}

    Although LLMs have achieved excellent performance in various tasks, it remains essential to explore how to train an LLM that can effectively perceive sentiments and evaluate performance. When training LLMs dedicated to AC tasks, researchers face the challenge of not only retaining the original abilities of models, such as instruction understanding and reasoning, but also strengthening new abilities, specifically sentiment perception and cognition. Additionally, it is crucial to ensure the integrity of values, facts, and logic within the LLMs.
    In terms of evaluation, previous studies conducted before the emergence of LLMs proposed various metrics to measure model performance, such as Accuracy, BLEU, and ROUGE. However, these metrics are insufficient to fully assess the ability of LLMs, e.g., generalization,  generation, perception, and cognitive. Establishing a comprehensive and effective evaluation standard remains an urgent problem to be solved.

    

    

\end{enumerate}